%% file: main.tex
\documentclass[]{worldbench}
\input{preamble}

\title{LargeAD: Large-Scale Cross-Sensor Data Pretraining for Autonomous Driving}

\author[]{Lingdong~Kong~\raisebox{0.2em}{\includegraphics[width=0.019\linewidth]{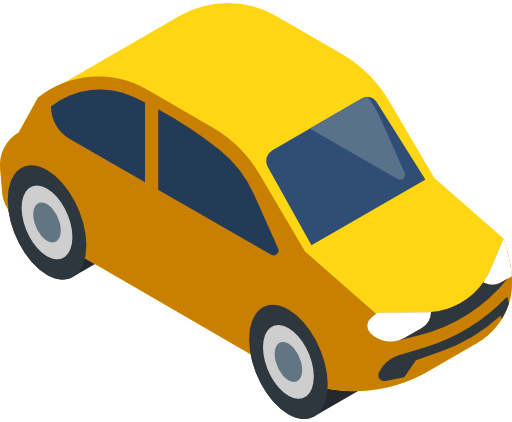}}~\raisebox{0.2em}{\includegraphics[width=0.019\linewidth]{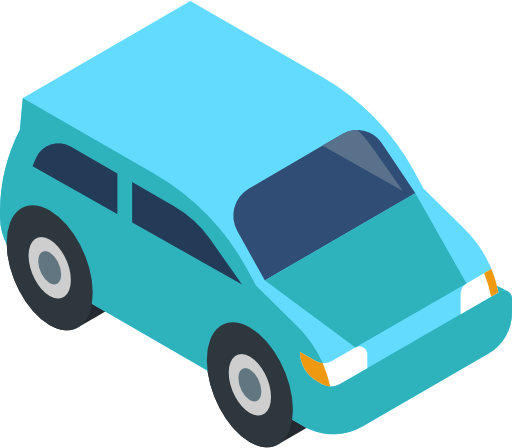}}}
\author[]{Xiang~Xu\raisebox{0.2em}{\includegraphics[width=0.019\linewidth]{figures/icons/car1.png}}}
\author[]{Youquan~Liu\raisebox{0.2em}{\includegraphics[width=0.019\linewidth]{figures/icons/car1.png}}}
\author[]{Jun~Cen}
\author[]{Runnan~Chen}
\author[]{Wenwei~Zhang}
\author[]{Liang~Pan}
\author[]{Kai~Chen}
\author[]{Ziwei~Liu~\raisebox{0.15em}{\includegraphics[width=0.017\linewidth]{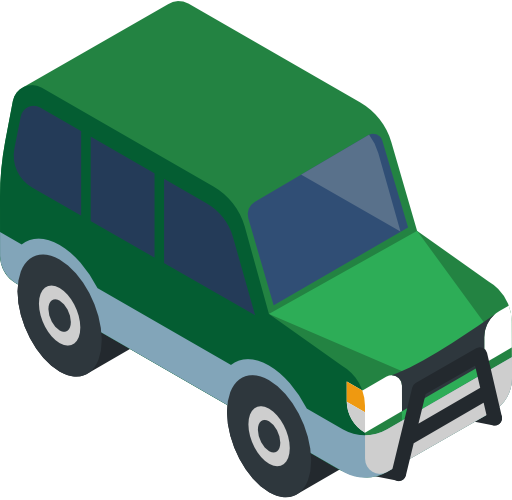}}}

\affiliation[]{
\raisebox{-0.1em}{\includegraphics[width=0.029\linewidth]{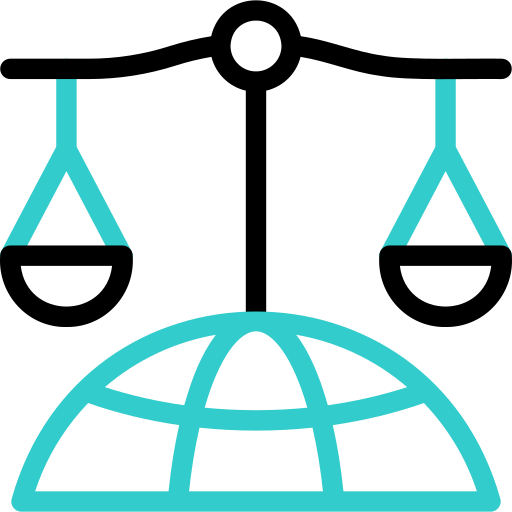}}~WorldBench Team
\\[1.2ex]
~\raisebox{-0.2em}{\includegraphics[width=0.032\linewidth]{figures/icons/car1.png}}~{\small \textbf{Equal Contributions}}
\quad
\raisebox{-0.2em}{\includegraphics[width=0.031\linewidth]{figures/icons/car2.png}}~{\small \textbf{Project Lead}}
\quad
\raisebox{-0.2em}{\includegraphics[width=0.028\linewidth]{figures/icons/car4.png}}~{\small \textbf{Corresponding Author}}
}

\abstract{
Recent advancements in vision foundation models (VFMs) have revolutionized visual perception in 2D, yet their potential for 3D scene understanding, particularly in autonomous driving applications, remains underexplored. In this paper, we introduce \textbf{LargeAD}, a versatile and scalable framework designed for large-scale 3D pretraining across diverse real-world driving datasets. Our framework leverages VFMs to extract semantically rich superpixels from 2D images, which are aligned with LiDAR point clouds to generate high-quality contrastive samples. This alignment facilitates cross-modal representation learning, enhancing the semantic consistency between 2D and 3D data. We introduce several key innovations: (i) VFM-driven superpixel generation for detailed semantic representation, (ii) a VFM-assisted contrastive learning strategy to align multimodal features, (iii) superpoint temporal consistency to maintain stable representations across time, and (iv) multi-source data pretraining to generalize across various LiDAR configurations. Our approach achieves substantial gains over state-of-the-art methods in linear probing and fine-tuning for LiDAR-based segmentation and object detection. Extensive experiments on 11 large-scale multi-sensor datasets highlight our superior performance, demonstrating adaptability, efficiency, and robustness in real-world autonomous driving scenarios.
}

\metadata[
\raisebox{-0.2em}{\includegraphics[width=0.025\linewidth]{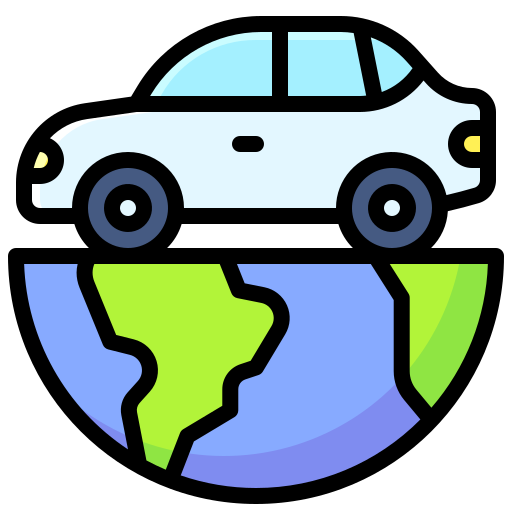}}~~Project Page]{\href{https://ldkong.com/LargeAD}{\texttt{https://ldkong.com/LargeAD}}
\\[-1.5ex]}

\begin{document}

\maketitle

\input{sections/1_intro}
\input{sections/2_related_work}
\input{sections/3_method}
\input{sections/4_new_approach}
\input{sections/5_experiments}
\input{sections/6_conclusion}

\bibliographystyle{ieeetr}
\bibliography{main}

\end{document}

%% file: preamble.tex
\definecolor{w_blue}{RGB}{52,204,204}
\definecolor{w_yellow}{RGB}{255,192,0}

\definecolor{forestgreen}{rgb}{0.133, 0.545, 0.133}
\definecolor{yellowyellow}{rgb}{0.133, 0.545, 0.133}

\definecolor{correct}{RGB}{173, 173, 173}
\definecolor{incorrect}{RGB}{192, 0, 0}

\usepackage{algorithmic}
\usepackage{algorithm}
\usepackage{amsmath}
\usepackage{amssymb}
\usepackage{amsthm}

\usepackage{booktabs}

\usepackage{caption}
\usepackage{color}
\usepackage{colortbl}

\usepackage{enumitem}

\usepackage{fontawesome}

\usepackage{graphicx}

\usepackage{hhline}
\usepackage{longtable}

\usepackage{makecell}
\usepackage{mathtools}
\usepackage{microtype}
\usepackage{multicol}
\usepackage{multirow}

\usepackage{pifont}

\usepackage{subcaption}

\usepackage[table,dvipsnames]{xcolor}
\usepackage{tabularx}
\usepackage{tikz}

\usepackage{wrapfig}

%% file: sections/1_intro.tex
\section{Introduction}
\label{sec:intro}

The emergence of large language models (LLMs) \cite{shoeyb2019megatronLM,hoffmann2022chinchilla,zhang2022opt,gpt4,chowdhery2022palm} has transformed natural language processing, paving the way for similar breakthroughs in computer vision through vision foundation models (VFMs) such as SAM \cite{kirillov2023sam}, X-Decoder \cite{zou2023xcoder}, and SEEM \cite{zou2023seem}. These models have demonstrated remarkable capabilities in extracting rich pixel-level semantics from 2D images. However, extending these advancements to the 3D domain remains a largely unexplored frontier. As autonomous driving applications increasingly rely on 3D data from LiDAR sensors, there is a growing need to transfer the success of VFMs in 2D vision to 3D scene understanding.

Accurate perception of LiDAR point clouds is crucial for safe autonomous driving and advanced driver-assistance systems \cite{sun2024lidarseg,triess2021survey,uecker2022analyzing,rizzoli2022survey,li2024place3d}. Traditional LiDAR point cloud models often depend on large annotated datasets, which are costly and time-consuming to create \cite{behley2021semanticKITTI,unal2022scribbleKITTI}. To alleviate this challenge, research has explored semi-supervised \cite{kong2023laserMix,li2023lim3d} and weakly-supervised \cite{unal2022scribbleKITTI,li2022coarse3D} methods. However, these approaches suffer from limited generalizability, especially when faced with diverse sensor configurations, \emph{e.g.}, different LiDAR beam numbers, camera placements, sampling rates, and potential sensor corruptions \cite{gao2021survey,xiao2023survey,triess2021survey,xie2024benchmarking,kong2023robodepth,hao2024mapbench,geiger2012automatic,hu2022investigating,li2024influence,chen2020novel}. This limitation poses a significant challenge for real-world scalability.

In response, we propose \textbf{LargeAD}, a novel and scalable framework for 3D scene understanding that leverages large-scale data pretraining across diverse sensors. Our approach builds on recent advances in cross-modal representation learning \cite{kirillov2023sam,zou2023xcoder,zhang2023openSeeD}, incorporating VFMs into the 3D domain to address several critical objectives: \emph{i)} utilizing raw point clouds as input to eliminate the need for costly labels, \emph{ii)} exploiting spatial and temporal cues from driving scenes for robust representation learning, and \emph{iii)} ensuring generalizability to downstream datasets beyond the pretraining data. By distilling the semantic knowledge encoded in VFMs, our methodology facilitates self-supervised learning on complex 3D point clouds, particularly for autonomous driving.

\begin{wrapfigure}{r}{0.58\textwidth}
    \begin{minipage}{\linewidth}
        \centering
        \vspace{-0.3cm}
        \includegraphics[width=\linewidth]{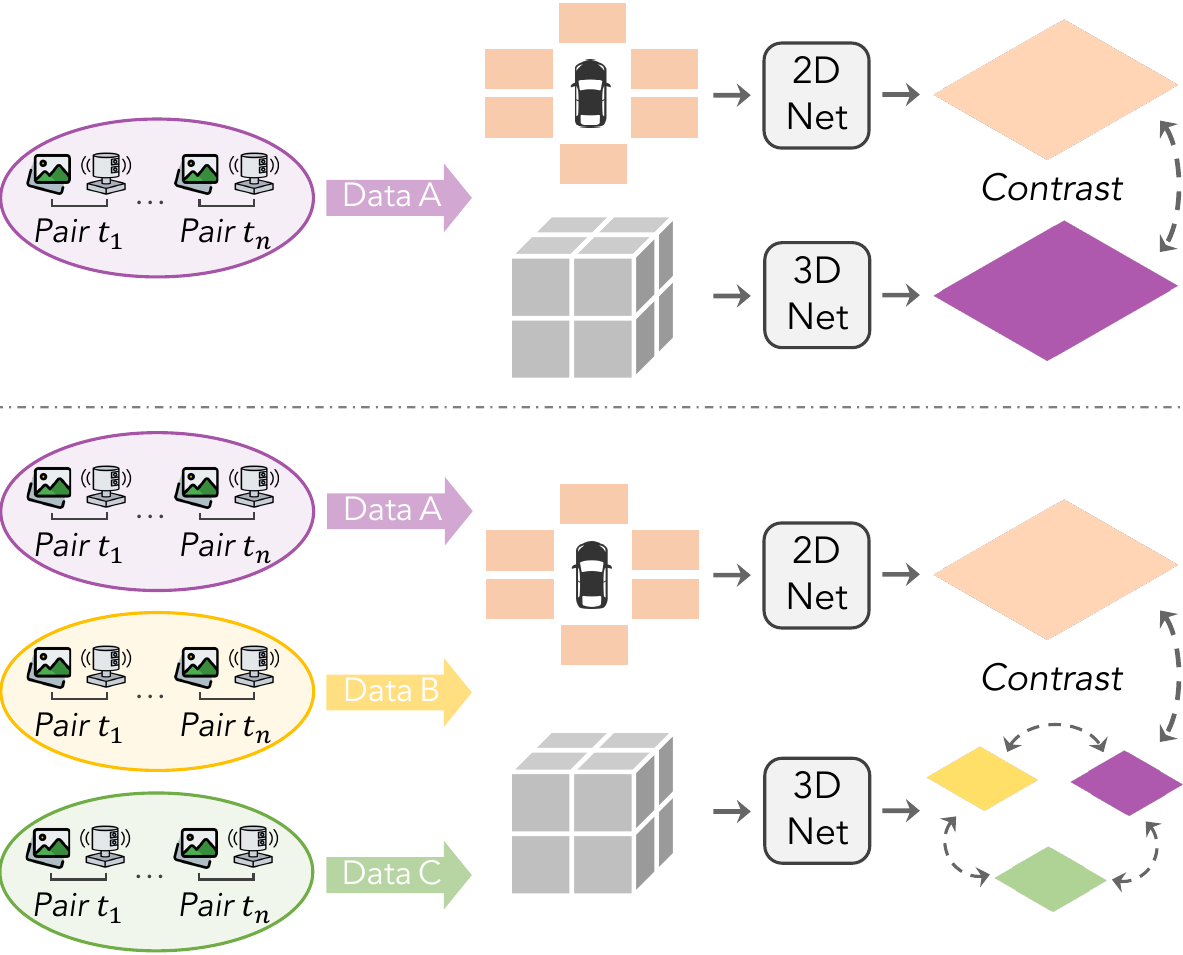}
        \vspace{-0.55cm}
        \caption{Comparisons between \emph{i)} the conventional image-to-LiDAR data pretraining frameworks \cite{sautier2022slidr,mahmoud2023st,liu2023seal} and \emph{ii)} our proposed large-scale cross-sensor data pretraining (\textbf{LargeAD}). Our approach combines heterogeneous data sources for representation learning, which achieves superior robustness and scalability. Different from previous work, our framework encourages representation learning across different datasets, which largely enhances the generalizability.}
        \label{fig:teaser_new}
    \end{minipage}
\end{wrapfigure}
A core innovation in our framework is the use of VFMs to generate semantically enriched superpixels from camera images, which are then aligned with LiDAR data to construct high-quality contrastive samples (see Figure~\ref{fig:teaser_new}). These semantic superpixels provide enhanced 2D-3D correspondences that capture object-level coherence, reducing the errors commonly associated with over-segmentation and ``self-conflict'' in contrastive learning \cite{liu2023seal}. This alignment significantly improves performance in downstream tasks, including 3D object detection and segmentation.

Furthermore, the proposed framework introduces several additional innovations. First, a VFM-assisted contrastive learning strategy aligns superpixels and superpoints within a unified embedding space, addressing the cross-modal discrepancies between image and LiDAR features. Second, a superpoint temporal consistency mechanism enhances the robustness of point cloud representations across time, mitigating errors from imperfect synchronization between LiDAR and camera sensors. Finally, our multi-source data pretraining strategy leverages diverse LiDAR datasets to build a generalized model capable of adapting to different sensor configurations, further boosting the scalability.

\begin{figure}[t]
    \begin{center}
    \includegraphics[width=\textwidth]{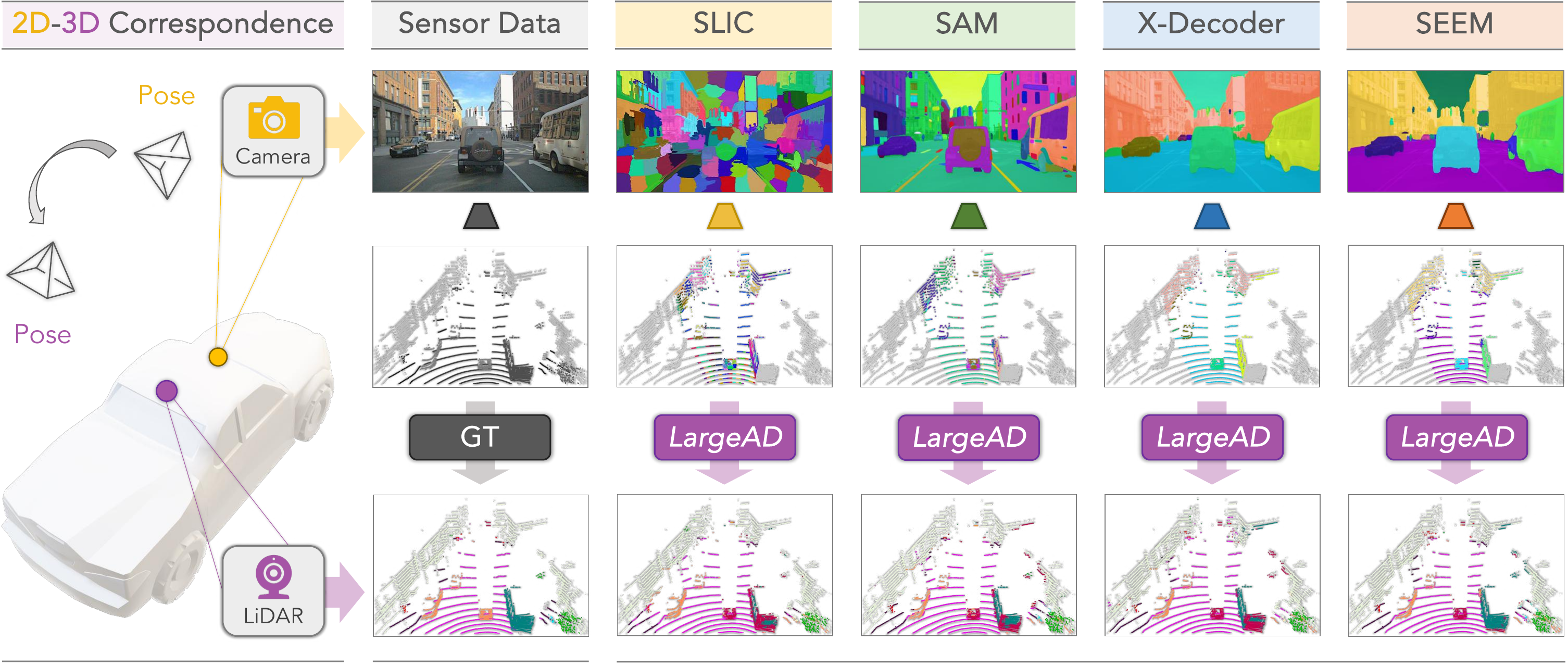}
    \end{center}
    \vspace{-0.1cm}
    \caption{Illustration of \textbf{image-to-LiDAR data pretraining} using \textit{i)} the heuristic SLIC algorithm \cite{achanta2012slic} and \textit{ii)} different vision foundation models (VFMs). Images in the first row are the superpixels generated by different methods, where each color represents one distinct segment. The LiDAR point clouds from the second row are the superpoints grouped by projecting superpixels to 3D using camera-LiDAR correspondence. The third row shows the linear probing results after data pretraining.}
\label{fig:teaser}
\end{figure}

As shown in Figure~\ref{fig:teaser}, compared to state-of-the-art methods such as SLidR \cite{sautier2022slidr} and ST-SLidR \cite{mahmoud2023st}, our framework introduces significant advancements: \emph{i)} the use of semantically rich superpixels to resolve the ``self-conflict'' issue in contrastive learning, \emph{ii)} the creation of high-quality contrastive samples, which lead to faster and more stable convergence, and \emph{iii)} reduced computational overhead due to the more efficient superpixel generation process.

%% file: sections/2_related_work.tex
\section{Related Work}
\label{sec:related_work}

In this section, we provide a thorough literature review of relevant works in the areas of LiDAR-based autonomous driving scene understanding (Section~\ref{sec:related_lidar}), popular vision foundation models (Section~\ref{sec:related_vfm}), different 3D representation learning techniques (Section~\ref{sec:related_pretrain}), and multi-dataset learning and utilization in the domain of 3D object detection and LiDAR semantic segmentation (Section~\ref{sec:related_multi}).

\subsection{LiDAR-Based Scene Understanding}
\label{sec:related_lidar}

For autonomous vehicles, accurate and dense 3D perception is crucial for safe navigation \cite{uecker2022analyzing,hu2021sensatUrban}. Researchers have developed various methods for point cloud segmentation, including those based on raw points \cite{hu2020randla,thomas2019kpconv,hu2022randla,zhang2023pids,puy23waffleiron}, range views \cite{milioto2019rangenet++,xu2020squeezesegv3,triess2020scan,zhao2021fidnet,cheng2022cenet,kong2023rethinking,xu2023frnet}, bird's eye views \cite{zhou2020polarNet,zhou2021panoptic}, voxels \cite{choy2019minkowski,zhu2021cylindrical,hong2021dsnet,hong20224dDSNet}, and multi-view fusion \cite{liong2020amvNet,tang2020searching,xu2021rpvnet,zhuang2021pmf,cheng2021af2S3Net,qiu2022gfNet}. Despite achieving significant advancements, these models typically rely on extensively annotated datasets, posing scalability issues \cite{gao2021survey}. To alleviate the annotation burden, recent studies have explored semi-supervised \cite{kong2023laserMix,li2023lim3d,kong2024lasermix2}, weakly-supervised \cite{unal2022scribbleKITTI,hu2022sqn,shi2022weak,liu2022weakly,li2022coarse3D}, and active learning \cite{liu2022less,hu2022inter,zhao2023semanticFlow} methods, as well as domain adaptation techniques \cite{jaritz2020xMUDA,jaritz2023xMUDA,kong2023conDA,peng2021sparse,saltori2020synth4D,michele2023saluda}. This work utilizes a self-supervised learning strategy, distilling knowledge from VFMs through camera-to-LiDAR associations, thereby eliminating the need for manual annotations during pretraining.

\subsection{Vision Foundation Models}
\label{sec:related_vfm}

The field of computer vision has been transformed by the development of vision foundation models (VFMs) that leverage vast amounts of training data \cite{radford2021clip,kirillov2023sam} and sophisticated self-supervised learning techniques \cite{caron2021dino,oquab2023dinov2}. Among these, the Segment Anything Model, or SAM \cite{kirillov2023sam}, has set a new benchmark in general-purpose image segmentation, showcasing impressive zero-shot transfer capabilities across a range of downstream tasks. Other notable VFMs, such as X-Decoder \cite{zou2023xcoder}, OpenSeeD \cite{zhang2023openSeeD}, SegGPT \cite{wang2023segGPT}, and SEEM \cite{zou2023seem}, have further demonstrated the versatility of these models in handling diverse image-related tasks. This work extends the utilization of VFMs to the domain of point cloud learning, capitalizing on their semantic understanding to enhance spatial and temporal cues in 3D representation learning.

\subsection{Representation Learning in 3D}
\label{sec:related_pretrain}

3D self-supervised learning has its roots in image-based techniques and typically focuses on object-centric point clouds \cite{sauder2019self,poursaeed2020shape,sanghi2020info3D,garg2022seRP,tran2022self} or indoor scenes \cite{chen20224dContrast,chen2023clip2Scene,lal2021coCoNets,li2022dpCo,yamada2022point} using pretext tasks \cite{doersch2015unsupervised,misra2016shuffle,noroozi2016jigsaw,zhang2016colorful,gidaris2018unsupervised}, contrastive learning \cite{chen2020simCLR,he2020moCo,chen2020moCoV2,henaff2020cpc,gansbeke2021unsupervised,chen2021moCoV3,henaff2021efficient}, or mask modeling \cite{xie2022simMIM,he2022mae,feichtenhofer2022mae,li2022scaling}. These methods often lack the necessary scale and diversity for outdoor driving scenes \cite{michele2021generative,boulch2022also,sautier2024bevcontrast}. Efforts such as PointContrast \cite{xie2020pointContrast}, DepthContrast \cite{zhang2021depthContrast}, and SegContrast \cite{nunes2022segcontrast} have pioneered contrastive objectives for small-scale point clouds. Recently, Sautier \emph{et al.} \cite{sautier2022slidr} introduced SLidR, the first method for image-to-LiDAR representation distillation in cross-modal self-supervised learning on large-scale point clouds. Mahmoud \emph{et al.} \cite{mahmoud2023st} further refined this approach with semantically tolerant contrastive constraints and a class-balancing loss. SuperFlow \cite{xu20244d} introduced a spatiotemporal consistency framework to efficiently capture dynamic cues across multiple timestamps. Our framework builds upon SLidR \cite{sautier2022slidr} by leveraging VFMs \cite{kirillov2023sam,zou2023xcoder,zou2023seem} to create a more effective cross-modal contrastive objective. We also introduce a superpoint temporal consistency regularization to enhance feature learning and robustness in diverse and dynamic real-world driving scenarios.

\subsection{Multi-Dataset Utilization}
\label{sec:related_multi}

Leveraging multiple datasets has emerged as a promising approach for improving the generalization of LiDAR-based models in autonomous driving \cite{kalluri2019universal}. Recent works like MDT3D \cite{fontez2023mdt3d} and MS3D++ \cite{tsai2023ms3d++} explored multi-source training for 3D object detection, while addressing challenges such as label space conflicts. Similarly, methods like COLA \cite{sanchez2022cola} and M3Net \cite{liu2024multi} utilized unified label spaces for semantic segmentation, demonstrating the advantages of multi-dataset learning. In this work, we extend these efforts by proposing a multi-source pretraining strategy that incorporates data from diverse LiDAR datasets, each with unique sensor setups and conditions \cite{caesar2020nuScenes,behley2019semanticKITTI,sun2020waymoOpen,li2024place3d,cocheteux2025uncertainty}. By aligning 2D and 3D features across different sources, our approach improves cross-modal consistency and robustness. To the best of our knowledge, this is the first comprehensive study on multi-source pretraining for image-to-LiDAR representation learning, allowing our framework to generalize more effectively across varied driving environments and sensor configurations.

%% file: sections/3_method.tex
\section{Image-to-LiDAR Data Pretraining}
\label{sec:approach}

In this section, we elaborate on the technical details of the image-to-LiDAR pretraining. We first present the preliminaries (Section~\ref{sec:preliminaries}), followed by the introduction of superpixel-driven contrastive learning (Section~\ref{sec:slidr}), and conclude with the description of the contrastive learning objective (Section~\ref{sec:objective}).

\begin{figure}[t]
\begin{center}
\includegraphics[width=\textwidth]{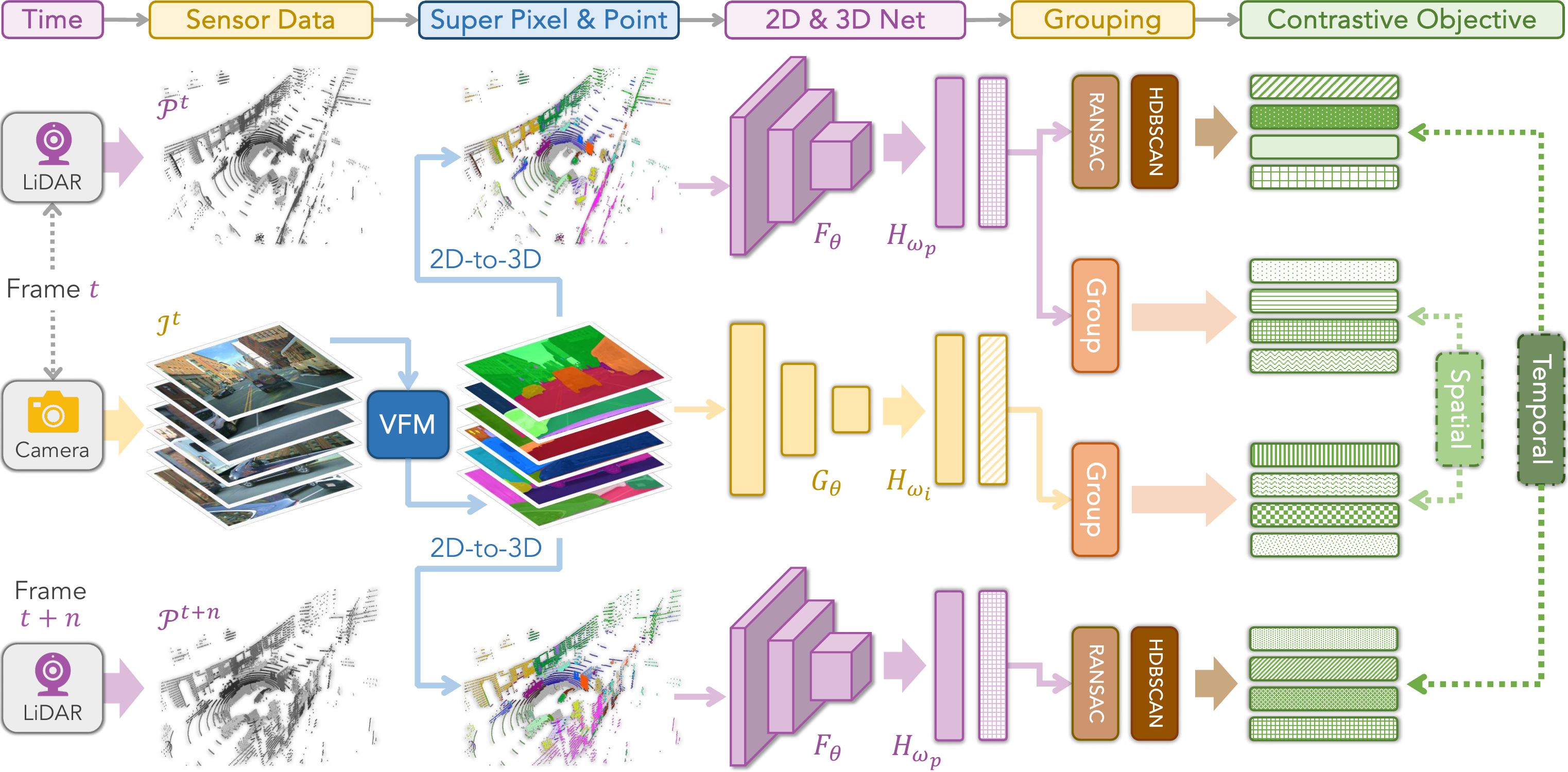}
\end{center}
\vspace{-0.1cm}
\caption{Overview of the \textbf{VFM-driven image-to-LiDAR contrastive learning} framework. Given a pair of LiDAR point cloud $\mathcal{P}^{t}$ and camera image $\mathcal{I}^{t}$ captured at timestamp $t$, along with another LiDAR point cloud $\mathcal{P}^{t+n}$ captured at timestamp $t+n$, we generate semantic superpixels using vision foundation models (VFMs). The corresponding superpoints are obtained by projecting image pixels onto the point cloud. Two key objectives are established: \emph{i)} spatial contrastive learning between paired LiDAR and camera features, and \emph{ii)} temporal consistency regularization between point segments from $\mathcal{P}^{t}$ and $\mathcal{P}^{t+n}$.}
\label{fig:framework}
\end{figure}

\subsection{Problem Formulation}
\label{sec:preliminaries}

Let us define a point cloud $\mathcal{P}=\{\mathbf{p}_i, \mathbf{e}_i \mid i = 1, \ldots, N\}$ consisting of $N$ points collected by a LiDAR sensor. Each point $\mathbf{p}_i \in \mathbb{R}^{3}$ denotes the 3D coordinates, while $\mathbf{e}_i \in \mathbb{R}^{L}$ represents its feature embedding, such as intensity, elongation, \emph{etc}. This work aims to transfer knowledge from a set of surround-view images $\mathcal{I}=\{\mathbf{I}_i \mid i = 1, \ldots, V\}$, captured by a total of $V$ synchronized RGB cameras, to the point cloud $\mathcal{P}$. Each image $\mathbf{I} \in \mathbb{R}^{3 \times H \times W}$ has a spatial resolution defined by height $H$ and width $W$.

Given that the LiDAR and camera sensors are assumed to be well-calibrated~\cite{ li2025unidrive,cattaneo2025cmrnext}, each LiDAR point $\mathbf{p}_i = (x_i, y_i, z_i)$ can be projected onto the image plane as a pixel $\hat{\mathbf{p}}_i = (u_i, v_i)$ using the following coordinate transformation:
\begin{align}
\label{eqn:point-pixel-pair}
[u_i, v_i, 1]^{\text{T}} = \frac{1}{z_{i}} \times \Gamma_{K} \times \Gamma_{c \leftarrow l} \times [x_i, y_i, z_i]^{\text{T}}~,
\end{align}
where $\Gamma_K$ denotes the camera intrinsic matrix, and $\Gamma_{c \leftarrow l}$ is the transformation matrix from the LiDAR to the camera coordinate system. 

Previous studies\cite{sautier2022slidr,mahmoud2023st} employ the unsupervised SLIC algorithm~\cite{achanta2012slic} to aggregate image regions with similar RGB attributes into a set of superpixels, denoted as $\Phi_\mathcal{S} = \{\mathbf{s}_m \mid m=1, \ldots, M\}$. Subsequently, the corresponding superpoint set $\Phi_{\mathcal{O}} = \{\mathbf{o}_m \mid m = 1, \ldots, M\}$ is derived using Eq.~\ref{eqn:point-pixel-pair}. To facilitate the knowledge transfer from images to the LiDAR domain, these methods \cite{sautier2022slidr,mahmoud2023st} usually conduct cross-modal contrastive learning between the representations of superpixels and superpoints.

\subsection{Superpixel-Driven Contrastive Learning}
\label{sec:slidr}

Earlier methods like PPKT~\cite{liu2021ppkt} align image pixels with corresponding LiDAR points through contrastive learning. However, PPKT~\cite{liu2021ppkt} tends to struggle with several limitations when applied to sparse point cloud data, such as misalignment due to viewpoint differences, inadequate local semantic modeling, imbalanced weighting of dense and sparse regions, and poor handling of false negatives. While it performs well in dense regions (\emph{e.g.}, near vehicles), its effectiveness drops significantly in sparse areas, limiting its overall generalization.

To overcome these issues, SLidR~\cite{sautier2022slidr} introduces a superpixel-driven distillation approach using the SLIC algorithm~\cite{achanta2012slic} to group similar pixels into coherent superpixels. By employing contrastive learning between superpixels in images and superpoints in LiDAR data, SLidR reduces alignment errors from sensor viewpoints and enhances local semantic consistency. Aggregating features at the superpixel and superpoint levels resolves the weighting imbalance present in PPKT~\cite{liu2021ppkt}, ensuring better treatment of both dense and sparse regions. Moreover, contrastive learning over larger regions helps reduce false negatives, leading to more robust image-to-LiDAR knowledge transfer.

\subsection{Contrastive Learning Objective}
\label{sec:objective}

Let $F_{\theta_{p}}: \mathbb{R}^{N \times (3 + L)} \to \mathbb{R}^{N \times C} $ represent a LiDAR point cloud encoder with trainable parameters $\theta_{p}$, which processes a point cloud $\mathcal{P}$ and outputs a $C$--dimensional feature for each point. Additionally, let $G_{\theta_{i}}: \mathbb{R}^{H \times W \times 3} \to \mathbb{R}^{\frac{H}{s} \times \frac{W}{s} \times E}$ be an image encoder with parameters $\theta_{i}$, initialized from 2D self-supervised pretrained models. To compute the superpixel-driven contrastive loss, we construct trainable projection heads $H_{\omega_p}$ and $H_{\omega_i}$ which map the 3D point features and 2D image features into the same $D$-dimensional embedding space. The point projection head $H_{\omega_p}: \mathbb{R}^{N \times C} \to \mathbb{R}^{N \times D}$ is a linear layer followed by $\ell_2$-normalization. The image projection head $H_{\omega_i}: \mathbb{R}^{\frac{H}{s} \times \frac{W}{s} \times E} \to \mathbb{R}^{H \times W \times D}$ consists of a convolution layer with a $1 \times 1$ kernel, followed by a fixed bi-linear interpolation layer in the spatial dimension, and outputs with $\ell_2$-normalization. 

The goal is to distill the 2D network’s knowledge into the 3D network, ensuring that each semantic superpoint feature closely correlates with its corresponding semantic superpixel feature. Specifically, the superpixels $\Phi_\mathcal{S}$ and superpoints $\Phi_\mathcal{O}$ are used to group pixel and point embedding features, respectively. An average pooling operation is applied to the grouped pixel and point embeddings to obtain superpixel embedding features $\mathbf{Q} \in \mathbb{R}^{M \times D}$ and superpoint embedding features $\mathbf{K} \in \mathbb{R}^{M \times D}$. The contrastive loss $\mathcal{L}^{\mathrm{slic}} = \mathcal{L}\left(\mathbf{Q}, \mathbf{K}\right)$ is then defined as follows:
\begin{align}
\label{eqn:loss_slic}
    \mathcal{L}^{\mathrm{slic}} = - \frac{1}{M} {\sum_{i=1}^{M} \log \left[\frac{e^{\left(\left<\mathbf{q}_i,\mathbf{k}_i\right>/\tau\right)}}{\sum_{j=1}^{M} e^{\left(\left<\mathbf{q}_i,\mathbf{k}_j\right>/\tau\right)}}\right]},
\end{align}
where $\left<\mathbf{q}_i,\mathbf{k}_j\right>$ represents the scalar product between superpoint and superpixel embedding features, measuring their similarity. $\tau$ is a temperature parameter used to scale the similarity scores.

%% file: sections/4_new_approach.tex
\section{LargeAD: A Scalable, Versatile, and Generalizable Framework}
\label{sec:large_ad}

In this section, we present the technical details of the proposed framework. We begin by introducing our VFM-based superpixel generation approach (Section~\ref{sec:vfm}), followed by the explanation of semantic spatial consistency learning (Section~\ref{sec:c2l}). Next, we discuss the instance superpoint temporal consistency (Section~\ref{sec:superpoint_temporal_consistency}), and conclude with the description of multi-source data pretraining (Section~\ref{sec:multi-source}), culminating with a summary of the overall framework (Section~\ref{sec:overall}).

\subsection{Superpixel Generation from Foundation Models}
\label{sec:vfm}
\begin{figure}[t]
\begin{center}
\includegraphics[width=\textwidth]{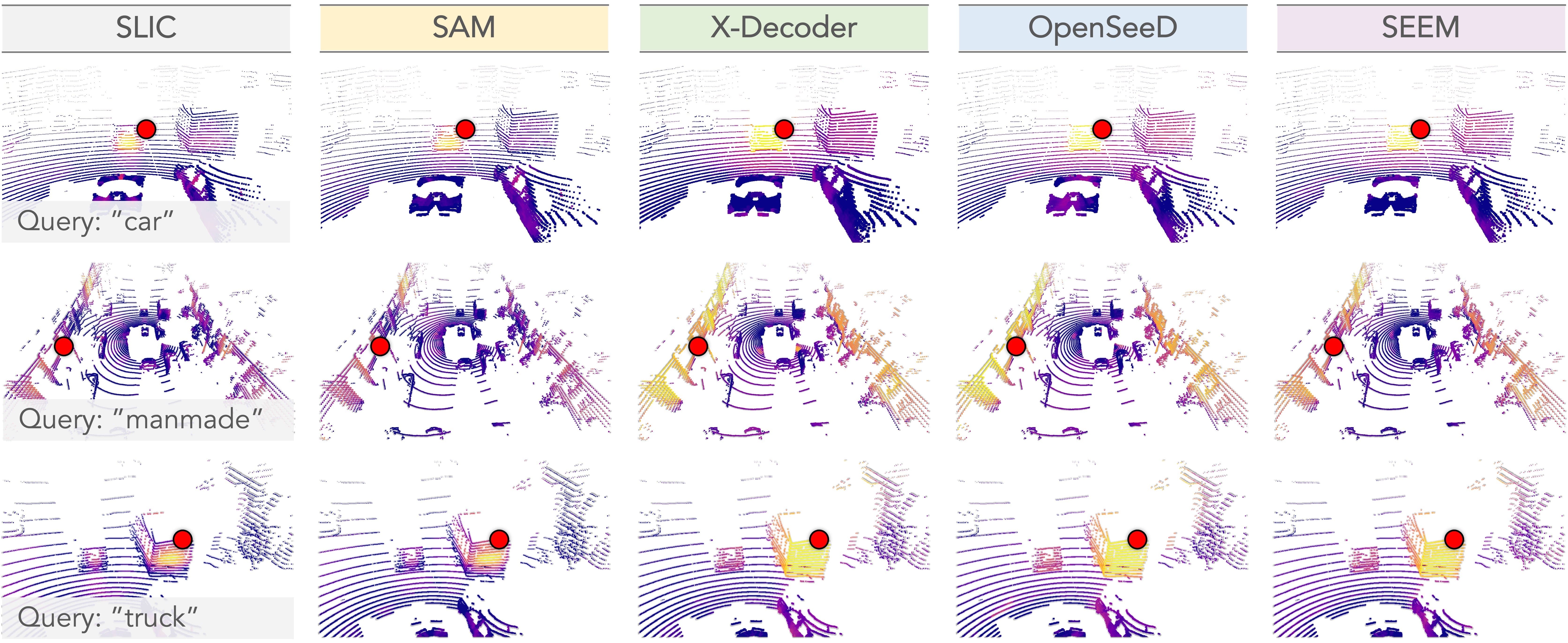}
\end{center}
\vspace{-0.2cm}
\caption{The \textbf{cosine similarity} between a query point (denoted as the \textcolor{red}{\textbf{red dot}}) and the feature learned with SLIC \cite{achanta2012slic} and different VFMs \cite{kirillov2023sam,zou2023xcoder,zhang2023openSeeD,zou2023seem}. The queried semantic classes from top to bottom examples are: ``car'', ``manmade'', and ``truck''. The color goes from \textcolor{violet}{\textbf{violet}} to \textcolor{Goldenrod}{\textbf{yellow}} denoting \textcolor{violet}{\textbf{low}} and \textcolor{Goldenrod}{\textbf{high}} similarity scores, respectively. Best viewed in colors.}
\label{fig:cosine}
\end{figure}

Previous works have utilized SLIC~\cite{achanta2012slic} to group visually similar image regions into superpixels. However, SLIC often over-segments semantically coherent areas (see Figure~\ref{fig:teaser}), which poses challenges for contrastive learning, particularly due to the ``self-conflict'' phenomenon. This occurs when semantically similar superpixels are incorrectly treated as negative samples\cite{wang2021understanding}. While \cite{mahmoud2023st} introduced a semantically tolerant loss to address this issue, the lack of high-level semantic understanding in SLIC exacerbates the difficulties in contrastive learning. To overcome these challenges, we generate semantic superpixels using vision foundation models (VFMs), which provide semantically rich superpixels and significantly improve the representation learning for both near and far points in the LiDAR point cloud (see Figure~\ref{fig:cosine}).

Instead of relying on low-level RGB features, our approach enhances superpixel generation by leveraging VFMs derived from large-scale pretrained image encoders \cite{kirillov2023sam, zou2023seem, zhang2023openSeeD, zou2023xcoder}. Unlike SLIC, VFMs capture high-level semantic information (as shown in Figure~\ref{fig:teaser}), allowing us to create more semantically meaningful superpixel sets, denoted as $\widehat{\Phi}_\mathcal{S} = \left\{ \left\{ \mathbf{s}_{m_v} \mid m_v=1, \ldots, M_v \right\} \mid M_v \ll M \right\}$. The generation process begins with the creation of semantic masks via prompts. By incorporating more abstract features, VFMs effectively address the ``self-conflict'' issue by grouping semantically similar regions more coherently, reducing the risk of misclassification during contrastive learning. As a result, the generated superpixels more accurately represent object semantics rather than just visual similarities. The corresponding superpoint set, $\widehat{\Phi}_\mathcal{O} = \left\{ \left\{ \mathbf{o}_{m_v} \mid m_v=1, \ldots, M_v \right\} \mid M_v \ll M \right\}$, is established using Eq.~\ref{eqn:point-pixel-pair}, ensuring proper alignment between the 2D image features and the 3D LiDAR point features.

Our VFM-assisted superpixels serve two primary purposes: first, they enhance the semantic richness of the generated superpixels, and second, they improve the alignment between 2D image features and the 3D LiDAR point cloud. By leveraging the high-level semantic features provided by VFMs, our approach effectively addresses issues like misalignment and feature inconsistency that often arise in traditional methods based on low-level RGB features. This enhanced semantic coherence between superpixels and superpoints reduces the occurrence of false negatives in contrastive learning. As a result, the improved feature alignment ensures that both superpixels and their corresponding superpoints more accurately reflect the underlying object semantics, ultimately leading to better performance in tasks such as 3D object detection and segmentation.

\subsection{Semantic Spatial Consistency Learning}
\label{sec:c2l}

Building upon the semantic superpixels generated by VFMs, as discussed in Sec.~\ref{sec:vfm}, we introduce a VFM-assisted contrastive learning framework that leverages these high-level visual features. The primary goal is to align superpixels with superpoints in a unified semantic space, ensuring that corresponding regions across different modalities are treated as positive pairs during training. By incorporating VFMs, this framework improves semantic consistency between images and LiDAR point clouds, addressing the alignment challenges often encountered in earlier methods. This approach enhances feature correspondence while reducing issues related to viewpoint variation and cross-modal discrepancies.

To implement this framework, we use the same trainable LiDAR point cloud encoder $F_{\theta_{p}}$ and frozen image encoder $G_{\theta_{i}}$ as described earlier, extracting features from the LiDAR point cloud and 2D images, respectively. For the contrastive loss, we employ the projection heads $H_{\omega_p}$ and $H_{\omega_i}$ from Section~\ref{sec:slidr}, projecting both point and image features into a shared $D$--dimensional embedding space. Unlike the low-level cues generated by SLIC, the superpixels produced by VFMs are enriched with semantic information, leading to more coherent and meaningful representations. 

To compute the VFM-assisted contrastive loss, we apply average pooling to the pixel and point embeddings grouped by the superpixel set $\widehat{\Phi}_\mathcal{S}$ and the corresponding superpoint set $\widehat{\Phi}_\mathcal{O}$. This process yields superpixel embeddings $\widehat{\mathbf{Q}} \in \mathbb{R}^{M_v \times D}$ and superpoint embeddings $\widehat{\mathbf{K}} \in \mathbb{R}^{M_v \times D}$. This contrastive loss $\mathcal{L}^{\mathrm{vfm}} = \mathcal{L}\left(\widehat{\mathbf{Q}}, \widehat{\mathbf{K}}\right)$ is formulated as follows:
\begin{align}
\label{eqn:loss_cross_modal}
    \mathcal{L}^{\mathrm{vfm}} = - \frac{1}{M_v} {\sum_{i=1}^{M_v} \log \left[\frac{e^{\left(\left< \hat{\mathbf{q}}_i, \hat{\mathbf{k}}_i \right>/\tau \right)}}{\sum_{j=1}^{M_v} e^{\left(\left< \hat{\mathbf{q}}_i, \hat{\mathbf{k}}_j \right>/\tau\right)}}\right]}.
\end{align}

The contrastive learning framework benefits from the rich semantic information provided by VFMs in several ways. First, these semantically enhanced superpixels help mitigate the ``self-conflict'' problem prevalent in existing approaches. Second, the high-quality contrastive samples generated by VFMs form a more coherent optimization landscape, leading to faster convergence compared to unsupervised superpixel generation methods. Finally, the use of superpixels from VFMs reduces the embedding length from hundreds (SLIC) to dozens, improving computational efficiency and accelerating the overall training process.

\subsection{Instance Superpoint Temporal Consistency}
\label{sec:superpoint_temporal_consistency}
\begin{wrapfigure}{r}{0.49\textwidth}
    \begin{minipage}{\linewidth}
        \centering
        \vspace{-0.55cm}
        \includegraphics[width=\linewidth]{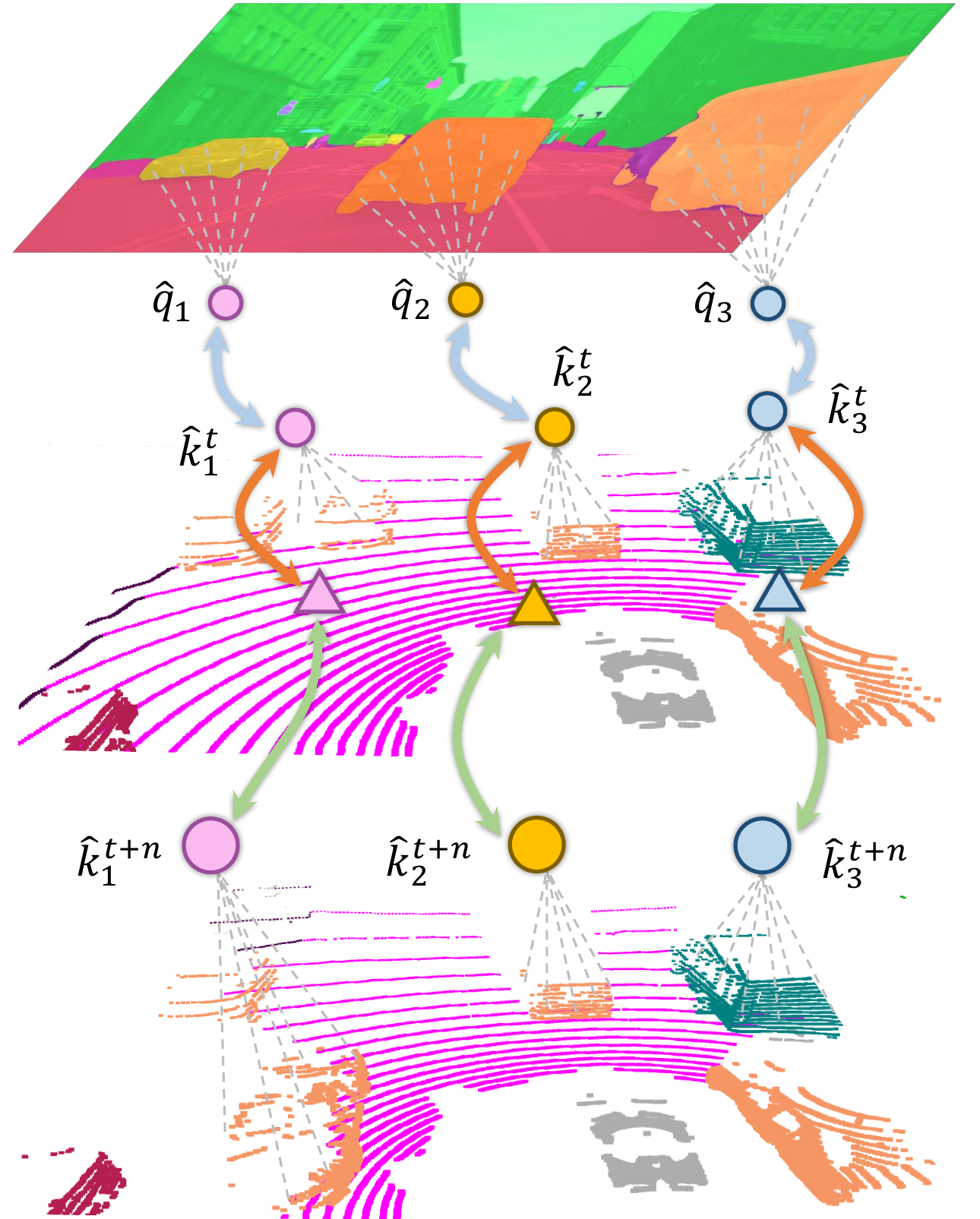}
        \vspace{-0.55cm}
        \caption{The \textbf{positive feature correspondences} of the contrastive objective in our proposed VFM-driven contrastive learning framework. The \emph{circles} and \emph{triangles} represent the instance-level and the point-level features, respectively.}
        \label{fig:temporal}
        \vspace{0.3cm}
    \end{minipage}
\end{wrapfigure}
In real-world deployments, perfectly synchronized LiDAR and camera data are often impractical, limiting scalability. To address this, we rely on accurate geometric information from point clouds to mitigate synchronization constraints.

\noindent\textbf{Implicit Geometric Clustering}. We first remove the ground plane points and select the non-ground points $\mathcal{G}^t$ from the LiDAR point cloud $\mathcal{P}^t$ at timestamp $t$ using the RANSAC algorithm \cite{foschler1981ransac}. We then group $\mathcal{G}^t$ into $M_k$ segments, $\mathcal{K}^t = \{\mathcal{K}_{1}^t,...,\mathcal{K}_{M_k}^t \}$, with the help of the HDBSCAN algorithm \cite{ester1996dbscan}. To map the segment views across different timestamps, we transform the LiDAR frames into a global coordinate frame, and then aggregate them. This gives the aggregated point cloud $\widetilde{\mathcal{P}} = \{\widetilde{\mathcal{P}}^t,... ,\widetilde{\mathcal{P}}^{t+n}\}$. Similarly, we generate non-ground plane $\widetilde{\mathcal{G}} = \{ \widetilde{\mathcal{G}}^t,... ,\widetilde{\mathcal{G}}^{t+n}\}$ from $\widetilde{\mathcal{P}}$ using RANSAC \cite{foschler1981ransac}. In the same manner as the single scan, we group $\widetilde{\mathcal{G}}$ to obtain $M_k$ segments $\widetilde{\mathcal{K}} =\{\widetilde{\mathcal{K}}_{1},...,\widetilde{\mathcal{K}}_{M_k} \}$. To generate the segment masks for all $n+1$ scans at $n$ consecutive timestamps, \emph{i.e.}, $\widetilde{\mathcal{K}} =\{\widetilde{\mathcal{K}}^{t},...,\widetilde{\mathcal{K}}^{t+n} \}$, we maintain the point index mapping from the aggregated point cloud $\widetilde{\mathcal{P}}$ to the $n+1$ individual scans.

\noindent\textbf{Superpoint Temporal Consistency}.
We leverage the clustered segments to compute the temporal consistency loss among related semantic superpoints\footnote{Here, we assume $n=1$, that is, the current and the next LiDAR frames, without loss of generalizability.}. Specifically, given a sampled temporal pair $\widetilde{\mathcal{P}}^t$ and $\widetilde{\mathcal{P}}^{t+1}$ and their corresponding segments $\widetilde{\mathcal{K}}^t$ and $\widetilde{\mathcal{K}}^{t+1}$, we compute the point-wise features $\widehat{\mathcal{F}}^t \in \mathbb{R}^{N \times D}$ and $\widehat{\mathcal{F}}^{t+1} \in \mathbb{R}^{N \times D}$ from the point projection head $H_{\omega_p}$. As for the target embedding, we split the point features $\widehat{\mathcal{F}}^t$ and $\widehat{\mathcal{F}}^{t+1}$ into $M_k$ groups by segments $\widetilde{\mathcal{K}}^t$ and $\widetilde{\mathcal{K}}^{t+1}$. Then, we apply an average pooling operation on $\widehat{\mathcal{F}}^{t+1}$ to get $M_k$ target mean feature vectors $\widehat{\mathcal{F}}^{t+1} = \{\widehat{\mathcal{F}}^{t+1}_1,\widehat{\mathcal{F}}^{t+1}_2,...,\widehat{\mathcal{F}}^{t+1}_{M_k}\}$, where $\widehat{\mathcal{F}}^{t+1}_{M_k} \in \mathbb{R}^{1 \times D}$. Let the split point feature $\widehat{\mathcal{F}}^t$ be $\widehat{\mathcal{F}}^{t} = \{\widehat{\mathcal{F}}^{t}_1, \widehat{\mathcal{F}}^{t}_2,...,\widehat{\mathcal{F}}^{t}_{M_k}\}$, where $\widehat{\mathcal{F}}^{t}_{M_k} \in \mathbb{R}^{k \times D}$ and $k$ is the number of points in the corresponding segment. We compute the temporal consistency loss ${\mathcal{L}}^{t \to {t+1}}$ to minimize the differences between the point features in the current frame (timestamp $t$) and the corresponding segment mean features from the next frame (timestamp $t+1$) as ${\mathcal{L}}^{t \to {t+1}}$ as follows:
\begin{equation}
\label{eqn:temporal_1}
    \mathcal{L}^{t \to {t+1}} = - \frac{1}{M_k} {\sum_{i=1}^{M_k} \log \left[\frac{e^{\left(\left<{\widehat{\mathbf{f}}}^t_i,{\widehat{\mathbf{f}}}^{t+1}_i\right>/\tau\right)}}{\sum_{j=1}^{M_k} e^{\left(\left<{\widehat{\mathbf{f}}}^t_i,{\widehat{\mathbf{f}}}^{t+1}_j\right>/\tau\right)}}\right]}.
\end{equation}
Since the target embedding for all points within a segment in the current frame serves as the mean segment representation from the next frame, this loss will force points from a segment to converge to a mean representation while separating from other segments, implicitly clustering together points from the same instance. Figure~\ref{fig:temporal} provides the positive feature correspondence in our contrastive learning framework. Furthermore, we swap $\widehat{\mathcal{F}}^{t}$ when generating the target mean embedding features to form a symmetric representation. In this way, the correspondence is encouraged from both $t \to {t+1}$ and ${t+1} \to t $, which leads to the following optimization objective: ${\mathcal{L}}^{\mathrm{tmp}} ={\mathcal{L}}^{t \to {t+1}}  + {\mathcal{L}}^{{t+1} \to t}$.

\noindent\textbf{Point to Segment Regularization}.
To pull close the LiDAR points belonging to the same instance at timestamp $t$, we minimize the distance between the point feature $\widehat{\mathcal{F}}^{t}$ and the corresponding mean cluster feature $\mathcal{C}^{t}$. To implement this, we leverage a max-pooling function to pool  $\widehat{\mathcal{F}}^{t}$ according to the segments to obtain $\mathcal{C}^{t} = \{\mathcal{C}^{t}_1, \mathcal{C}^{t}_2,...,\mathcal{C}^{t}_{M_k}\}$, where $\mathcal{C}^{t}_{M_k} \in \mathbb{R}^{1 \times D}$. The point-to-segment regularization is thus achieved via the ${\mathcal{L}}^{\mathrm{p2s}}$ loss function as follows:
\begin{equation}
    \mathcal{L}^{\mathrm{p2s}} = - \frac{1}{M_k N_k} {\sum_{i=1}^{M_k} \sum_{a=1}^{N_k}\log \left[\frac{e^{\left(\left<{\mathbf{c}}^t_i,{\widehat{\mathbf{f}}}^t_{i,a}\right>/\tau\right)}}{\sum_{j=1}^{M_k} e^{\left(\left<{\mathbf{c}}^t_i,{\widehat{\mathbf{f}}}^{t}_{j,a}\right>/\tau\right)}}\right]},
\label{eqn:p2s}
\end{equation}
where $N_k$ represents the number of points within the corresponding segment. The final optimization objective is to minimize the aforementioned semantic spatial consistency loss $\mathcal{L}^{\mathrm{vfm}}$, temporal consistency loss $\mathcal{L}^{\mathrm{tmp}}$, and the point-to-segment regularization loss $\mathcal{L}^{\mathrm{p2s}}$.

Our semantic superpoint temporal consistency leverages accurate geometric information from point clouds, ensuring consistent representations across timestamps. This approach remains robust when 2D-3D correspondence between LiDAR and cameras is unreliable, mitigating errors from calibration or synchronization issues. The point-to-segment regularization further improves spatial aggregation, enhancing the model’s ability to distinguish instances, \emph{e.g.}, ``car'' and ``truck''. Our experimental results confirm that these regularization strategies not only improve representation learning but also maintain effectiveness under sensor perturbations.

\subsection{Multi-Source Data Pretraining}
\label{sec:multi-source}

Previous works~\cite{sautier2022slidr,mahmoud2023st} predominantly focus on pretraining models with single-source LiDAR datasets. This approach limits their generalization capability when applied to out-of-source tasks, as different LiDAR datasets often exhibit diverse characteristics, such as varying sparsity, intensity ranges, and other sensor-specific attributes. To overcome these limitations, we propose a multi-source data pretraining strategy that integrates diverse datasets, improving the robustness of feature representations. This strategy enhances the model’s adaptability to different LiDAR sensors and boosts its generalization performance across domains.

\noindent\textbf{Multi-Source Contrastive Learning}.
Consider multiple LiDAR datasets $\{\{\mathcal{P}^{(i)}; \mathcal{I}^{(i)}\} \mid i = 1, \ldots, S\}$ from $S$ distinct sources\footnote{Here, to pursue better simplicity, we omit the timestamp $t$ and use $(i)$ to represent different sources.}. Our LiDAR point cloud network $F_{\theta_p}$ is designed to perform consistently across all sensors. However, significant discrepancies exist in the feature distributions of these datasets. For instance, intensity values in nuScenes \cite{caesar2020nuScenes}, range from $0$ to $255$; whereas the intensity values in SemanticKITTI ~\cite{behley2019semanticKITTI} range from $0$ to $1$. These differences complicate the learning process when using shared model weights across datasets.

To address these domain-specific variations, we first normalize the feature embeddings for each data source. For each dataset, we compute the mean $\epsilon^{(i)}$ and variance $\sigma^{(i)}$ of the feature distributions, then normalize the feature embeddings as follows:
\begin{equation}
\label{eqn:feature_normalization}
\hat{e}^{(i)}_j = \frac{e^{(i)}_j - \epsilon^{(i)}}{\sigma^{(i)}} .
\end{equation}
This normalization ensures consistent feature representation across datasets, minimizing the impact of differing distribution characteristics. After normalization, the feature embeddings are fed into the network $F_{\theta_p}$, which generates point features that are grouped into superpoint embeddings, $\{\mathbf{K}^{(i)} \mid i = 1, \dots, S\}$, for each domain. 

To improve the model’s generalizability across datasets, we employ a cross-dataset pretraining contrastive loss $\mathcal{L}^{\mathrm{cdp}}$, which encourages the model to learn shared representations across data sources while preserving the unique characteristics of each domain. This loss is defined as follows:
\begin{align}
\label{eqn:loss_cross_source}
\mathcal{L}^{\mathrm{cdp}} = - \frac{1}{M} {\sum_{i=1}^{M} \log \left[\frac{e^{\left(\left<\mathbf{k}^{(m)}_i,\mathbf{k}^{(n)}_i\right>/\tau\right)}}{\sum_{j=1}^{M} e^{\left(\left<\mathbf{k}^{(m)}_i,\mathbf{k}^{(n)}_j\right>/\tau\right)}}\right]}~.
\end{align}
Here, this loss ensures that superpoint embeddings from the same source are more similar, while simultaneously maintaining sufficient separation between superpoints from different sources. This contrastive objective strengthens the model’s ability to handle data from multiple domains and encourages the development of shared, yet adaptable, feature representations.

The multi-source data pretraining leverages a variety of data sources to create a more resilient and flexible model. By addressing the substantial distributional differences between domains, feature normalization ensures consistency across diverse datasets, facilitating a more unified representation space. The cross-source contrastive loss further enhances adaptability, enabling it to transfer knowledge across multiple domains while maintaining domain-specific distinctions. As we will show in Section~\ref{sec:experiment}, this strategy results in improved generalization across different LiDAR sensors, leading to stronger performance on out-of-source downstream tasks.

\input{tables/pre_nuscenes}

\subsection{Overall Framework}
\label{sec:overall}

Our framework integrates several innovative components to achieve scalable and robust 3D scene understanding. A key element is the use of VFMs to generate semantically enriched superpixels, addressing issues like over-segmentation and self-conflict in traditional methods. This enables better alignment between 2D image features and 3D LiDAR data, enhancing overall representation learning.

Our method incorporates the VFM-assisted contrastive loss $\mathcal{L}^{\mathrm{vfm}}$, ensuring semantic coherence between superpixels and superpoints, while the temporal consistency loss $\mathcal{L}^{\mathrm{tmp}}$ maintains stable point representations across frames. The point-to-segment regularization loss $\mathcal{L}^{\mathrm{p2s}}$ further improves spatial coherence within segments. Finally, the cross-dataset pretraining loss $\mathcal{L}^{\mathrm{cdp}}$ addresses domain-specific variations, enhancing the model’s ability to generalize across different LiDAR sensors.
To sum up, the overall optimization objective of our framework is to minimize:
\begin{align}
\label{eqn:overall}
\mathcal{L} = \mathcal{L}^{\mathrm{vfm}} + \mathcal{L}^{\mathrm{tmp}} + \mathcal{L}^{\mathrm{p2s}} + \mathcal{L}^{\mathrm{cdp}}~.
\end{align}
Together, these objectives create a robust and versatile framework, ensuring superior performance across various tasks and domains while maintaining scalability and adaptability in real-world applications.

%% file: tables/pre_nuscenes.tex
\begin{table*}[t]
    \centering
    \caption{Benchmarking data pretraining methods pretrained on \emph{nuScenes} \cite{fong2022panoptic-nuScenes} and fine-tuned on \emph{nuScenes} \cite{caesar2020nuScenes}, \emph{SemanticKITTI} \cite{behley2019semanticKITTI}, and \emph{Waymo} \cite{sun2020waymoOpen}. The evaluations are conducted on the official validation split of each dataset. Group (a): random initialization. Group (b): single-modality data pretraining. Group (c): image-to-LiDAR data pretraining by distilling ResNet \cite{he2016residual}. Groups (d), (e), and (f): image-to-LiDAR data pretraining by distilling ViT \cite{dosovitskiy2020vit} (Small, Medium, Large) from DINOv2 \cite{oquab2023dinov2}. \textbf{LP} denotes linear probing with frozen backbones. All mIoU scores are given in percentage (\%). The \textbf{best} and \underline{second best} scores under each group are highlighted in \textbf{bold} and \underline{underline}.}
    \vspace{-0.1cm}
\label{tab:pre_nuscenes}
\resizebox{\linewidth}{!}{
    \begin{tabular}{c|r|r|cc|cccccc|c|c}
    \toprule
    \multirow{2}{*}{\textbf{\#}} & \multirow{2}{*}{\textbf{Method}} & \multirow{2}{*}{\textbf{Venue}} & \textbf{Backbone} & \textbf{Backbone} & \multicolumn{6}{c}{\textbf{nuScenes}} \vline & \textbf{KITTI} & \textbf{Waymo}
    \\
    & & & (2D) & (3D) & {\textbf{LP}} & {\textbf{1\%}} & {\textbf{5\%}} & {\textbf{10\%}} & {\textbf{25\%}} & {\textbf{Full}} & {\textbf{1\%}} & {\textbf{1\%}}
    \\\midrule\midrule
    (a) & \cellcolor{yellow!8}Random & \cellcolor{yellow!8}- & \cellcolor{yellow!8}- & \cellcolor{yellow!8}MinkU-34 & \cellcolor{yellow!8}8.10 & \cellcolor{yellow!8}30.30 & \cellcolor{yellow!8}47.84 & \cellcolor{yellow!8}56.15 & \cellcolor{yellow!8}65.48 & \cellcolor{yellow!8}74.66 & \cellcolor{yellow!8}39.50 & \cellcolor{yellow!8}39.41
    \\\midrule
    \multirow{4}{*}{(b)} & PointContrast \cite{xie2020pointContrast} & ECCV'20 & - & MinkU-34 & \underline{21.90} & 32.50 & - & - & - & - & \underline{41.10} & -
    \\
    & DepthContrast \cite{zhang2021depthContrast} & ICCV'21 & - & MinkU-34 & \textbf{22.10} & 31.70 & - & - & - & - & \textbf{41.50} & -
    \\
    & ALSO \cite{boulch2022also} & CVPR'23 & - & MinkU-34 & - & \underline{37.70} & - & \underline{59.40} & - & \underline{72.00} & - & -
    \\
    & BEVContrast \cite{sautier2024bevcontrast} & 3DV'24 & - & MinkU-34 & - & \textbf{38.30} & - & \textbf{59.60} & - & \textbf{72.30} & - & -
    \\\midrule
    \multirow{8}{*}{(c)} & PPKT \cite{liu2021ppkt} & arXiv'21 & ResNet-50 & MinkU-34 & 35.90 & 37.80 & 53.74 & 60.25 & 67.14 & 74.52 & 44.00 & \underline{47.60}
    \\
    & SLidR \cite{sautier2022slidr} & CVPR'22 & ResNet-50 & MinkU-34 & 38.80 & 38.30 & 52.49 & 59.84 & 66.91 & 74.79 & 44.60 & 47.12
    \\
    & ST-SLidR \cite{mahmoud2023st} & CVPR'23 & ResNet-50 & MinkU-34 & 40.48 & 40.75 & 54.69 & 60.75 & 67.70 & 75.14 & 44.72 & 44.93
    \\
    & TriCC \cite{pang2023tricc} & CVPR'23 & ResNet-50 & MinkU-34 & 38.00 & 41.20 & 54.10 & 60.40 & 67.60 & 75.60 & 45.90 & -
    \\
    & Seal \cite{liu2023seal} & NeurIPS'23 & ResNet-50 & MinkU-34 & 44.95 & 45.84 & 55.64 & 62.97 & 68.41 & 75.60 & 46.63 & \underline{49.34}
    \\
    & CSC \cite{chen2024csc} & CVPR'24 & ResNet-50 & MinkU-34 & \underline{46.00} & \underline{47.00} & \textbf{57.00} & \underline{63.30} & 68.60 & 75.70 & 47.20 & -
    \\
    & HVDistill \cite{zhang2024hvdistill} & IJCV'24 & ResNet-50 & MinkU-34 & 39.50 & 42.70 & 56.60 & 62.90 & \underline{69.30} & \textbf{76.60} & \textbf{49.70} & -
    \\
    & \cellcolor{violet!7}\textcolor{violet!66}{\textbf{LargeAD}} & \cellcolor{violet!7}\textbf{Ours} & \cellcolor{violet!7}ResNet-50 & \cellcolor{violet!7}MinkU-34 & \cellcolor{violet!7}\textbf{46.13} & \cellcolor{violet!7}\textbf{47.08} & \cellcolor{violet!7}\underline{56.90} & \cellcolor{violet!7}\textbf{63.74} & \cellcolor{violet!7}\textbf{69.34} & \cellcolor{violet!7}\underline{76.03} & \cellcolor{violet!7}\underline{49.55} & \cellcolor{violet!7}\textbf{50.29}
    \\\midrule
    \multirow{5}{*}{(d)} & PPKT~\cite{liu2021ppkt} & arXiv'21 & ViT-S & MinkU-34 & 38.60 & 40.60 & 52.06 & 59.99 & 65.76 & 73.97 & 43.25 & 47.44
    \\
    & SLidR~\cite{sautier2022slidr} & CVPR'22 & ViT-S & MinkU-34 & 44.70 & 41.16 & 53.65 & 61.47 & 66.71 & 74.20 & 44.67 & 47.57
    \\
    & Seal~\cite{liu2023seal} & NeurIPS'23 & ViT-S & MinkU-34 & \underline{45.16} & \underline{44.27} & \underline{55.13} & \underline{62.46} & \underline{67.64} & \underline{75.58} & \underline{46.51} & \underline{48.67}
    \\
    & ScaLR \cite{puy2024scalr} & CVPR'24 & ViT-S & MinkU-34 & 42.40 & 40.50 & - & - & - & - & - & -
    \\
    & \cellcolor{violet!7}\textcolor{violet!66}{\textbf{LargeAD}} & \cellcolor{violet!7}\textbf{Ours} & \cellcolor{violet!7}ViT-S & \cellcolor{violet!7}MinkU-34 & \cellcolor{violet!7}\textbf{46.58} & \cellcolor{violet!7}\textbf{46.78} & \cellcolor{violet!7}\textbf{57.33} & \cellcolor{violet!7}\textbf{63.85} & \cellcolor{violet!7}\textbf{68.66} & \cellcolor{violet!7}\textbf{75.75} & \cellcolor{violet!7}\textbf{50.07} & \cellcolor{violet!7}\textbf{50.83}
    \\\midrule
    \multirow{4}{*}{(e)} & PPKT \cite{liu2021ppkt} & arXiv'21 & ViT-B & MinkU-34 & 39.95 & 40.91 & 53.21 & 60.87 & 66.22 & 74.07 & 44.09 & 47.57
    \\
    & SLidR \cite{sautier2022slidr} & CVPR'22 & ViT-B & MinkU-34 & 45.35 & 41.64 & 55.83 & 62.68 & 67.61 & 74.98 & 45.50 & 48.32
    \\
    & Seal \cite{liu2023seal} & NeurIPS'23 & ViT-B & MinkU-34 & \underline{46.59} & \underline{45.98} & \underline{57.15} & \underline{62.79} & \underline{68.18} & \underline{75.41} & \underline{47.24} & \underline{48.91}
    \\
    & \cellcolor{violet!7}\textcolor{violet!66}{\textbf{LargeAD}} & \cellcolor{violet!7}\textbf{Ours} & \cellcolor{violet!7}ViT-B & \cellcolor{violet!7}MinkU-34 & \cellcolor{violet!7}\textbf{47.84} & \cellcolor{violet!7}\textbf{48.37} & \cellcolor{violet!7}\textbf{59.36} & \cellcolor{violet!7}\textbf{64.11} & \cellcolor{violet!7}\textbf{69.02} & \cellcolor{violet!7}\textbf{75.91} & \cellcolor{violet!7}\textbf{50.68} & \cellcolor{violet!7}\textbf{51.52}
    \\\midrule
    \multirow{4}{*}{(f)} & PPKT \cite{liu2021ppkt} & arXiv'21 & ViT-L & MinkU-34 & 41.57 & 42.05 & 55.75 & 61.26 & 66.88 & 74.33 & 45.87 & 47.82
    \\
    & SLidR \cite{sautier2022slidr} & CVPR'22 & ViT-L & MinkU-34 & 45.70 & 42.77 & 57.45 & 63.20 & 68.13 & 75.51 & 47.01 & 48.60
    \\
    & Seal \cite{liu2023seal} & NeurIPS'23 & ViT-L & MinkU-34 & \underline{46.81} & \underline{46.27} & \underline{58.14} & \underline{63.27} & \underline{68.67} & \underline{75.66} & \underline{47.55} & \underline{50.02}
    \\
    & \cellcolor{violet!7}\textcolor{violet!66}{\textbf{LargeAD}} & \cellcolor{violet!7}\textbf{Ours} & \cellcolor{violet!7}ViT-L & \cellcolor{violet!7}MinkU-34 & \cellcolor{violet!7}\textbf{48.71} & \cellcolor{violet!7}\textbf{49.21} & \cellcolor{violet!7}\textbf{60.37} & \cellcolor{violet!7}\textbf{64.82} & \cellcolor{violet!7}\textbf{69.85} & \cellcolor{violet!7}\textbf{75.94} & \cellcolor{violet!7}\textbf{51.68} & \cellcolor{violet!7}\textbf{52.68}
    \\\bottomrule
\end{tabular}
}
\end{table*}

%% file: sections/5_experiments.tex
\section{Experiments}
\label{sec:experiment}
In this section, we conduct extensive experiments to validate the effectiveness of the proposed \textbf{LargeAD} framework. We first introduce the datasets (Section~\ref{sec:exp_datasets}) and describe the implementation details (Section~\ref{sec:exp_implementation}). Next, we present a comparative study, demonstrating the superior performance of our framework compared to state-of-the-art methods across multiple tasks and datasets (Section~\ref{sec:comparative}). Finally, we provide an in-depth ablation study to analyze the impact of different components and data sources on the performance (Section~\ref{sec:ablation}).

\subsection{Datasets}
\label{sec:exp_datasets}
We evaluate the effectiveness of our approach using \textbf{\emph{eleven}} diverse datasets. The first group includes large-scale real-world LiDAR datasets: $^1$\textbf{\emph{nuScenes}} \cite{caesar2020nuScenes,fong2022panoptic-nuScenes}, $^2$\textbf{\emph{SemanticKITTI}} \cite{behley2019semanticKITTI}, and $^3$\textbf{\emph{Waymo Open}} \cite{sun2020waymoOpen}. These datasets capture urban driving scenes, with nuScenes utilizing a Velodyne HDL32E sensor, while SemanticKITTI and Waymo Open use 64-beam LiDARs. We also include $^4$\textbf{\emph{ScribbleKITTI}} \cite{unal2022scribbleKITTI}, which shares data with SemanticKITTI but offers weak annotations in the form of line scribbles. For off-road scenarios, we consider $^5$\textbf{\emph{RELLIS-3D}} \cite{jiang2021rellis3D}, which contains multimodal data from a campus environment, and $^6$\textbf{\emph{SemanticPOSS}} \cite{pan2020semanticPOSS}, a smaller dataset with a focus on dynamic objects. Additionally, we incorporate $^7$\textbf{\emph{SemanticSTF}} \cite{xiao2023semanticSTF}, which provides LiDAR scans collected in adverse weather conditions. Three synthetic datasets are also utilized: $^8$\textbf{\emph{SynLiDAR}} \cite{xiao2022synLiDAR}, $^9$\textbf{\emph{Synth4D}} \cite{saltori2020synth4D}, and $^{10}$\textbf{\emph{DAPS-3D}} \cite{klokov2023daps3D}, all generated using simulators to provide diverse driving environments and scenarios. Finally, we assess robustness on $^{11}$\textbf{\emph{nuScenes-C}} \cite{kong2023robo3D}, a benchmark from the Robo3D challenge featuring eight out-of-distribution corruptions common in real-world driving. For further details, we refer readers to the respective publications associated with each dataset.

\input{tables/pre_semantickitti}
\input{tables/multiple_datasets}
\input{tables/robustness}

\begin{figure*}[t]
    \begin{center}
    \includegraphics[width=\textwidth]{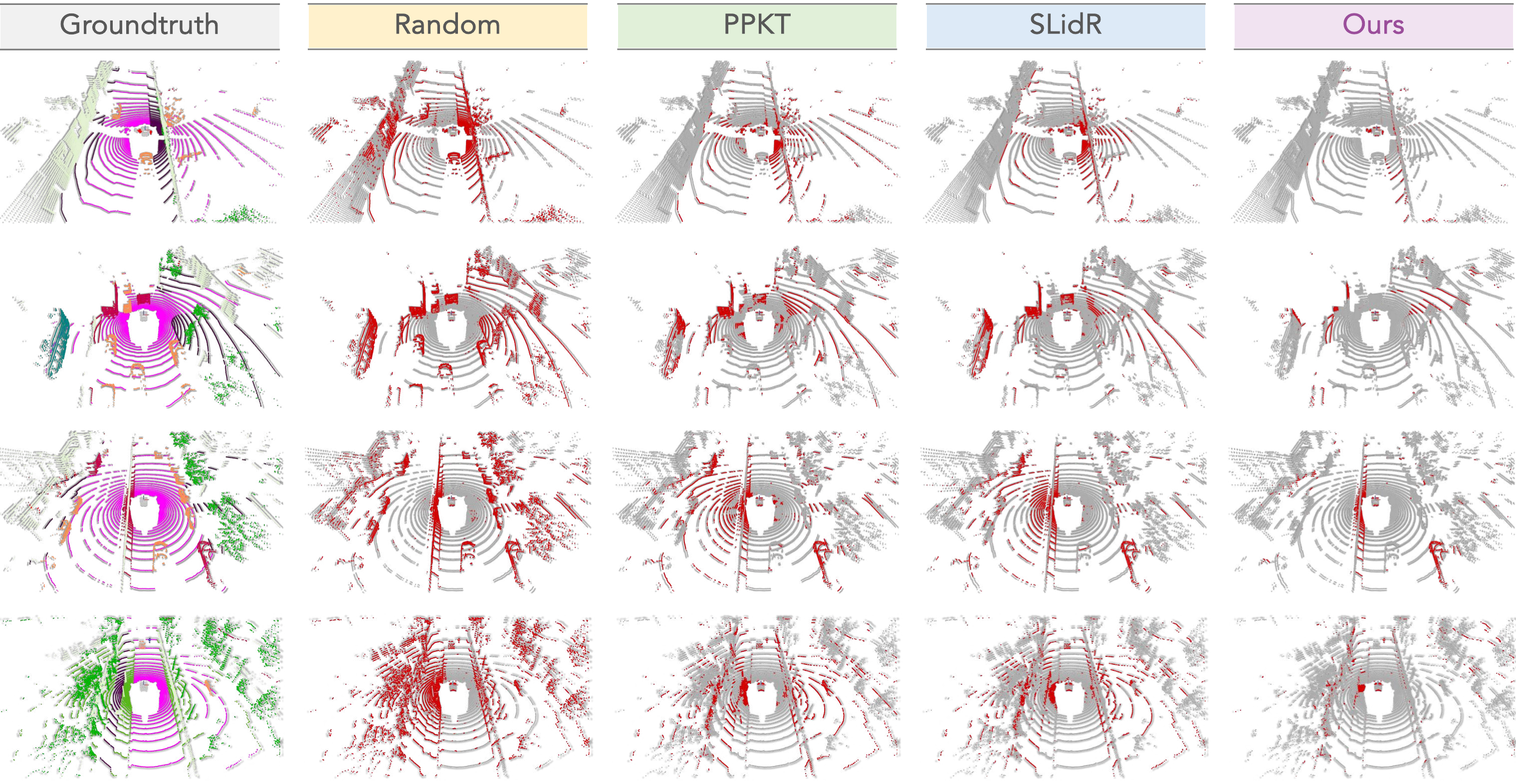}
    \end{center}
    \vspace{-0.2cm}
    \caption{The \textbf{qualitative results} of random initialization and different image-to-LiDAR data pretraining approaches \cite{liu2021ppkt,sautier2022slidr} pretrained and fine-tuned (with $1\%$ labeled data) on \emph{nuScenes} \cite{fong2022panoptic-nuScenes}. To highlight the differences, the \textbf{\textcolor{correct}{correct}} / \textbf{\textcolor{incorrect}{incorrect}} predictions are painted in \textbf{\textcolor{correct}{gray}} / \textbf{\textcolor{incorrect}{red}}, respectively. Best viewed in colors and zoomed-in for additional details.}
\label{fig:qualitative}
\end{figure*}

\begin{figure*}[t]
    \begin{center}
    \includegraphics[width=\textwidth]{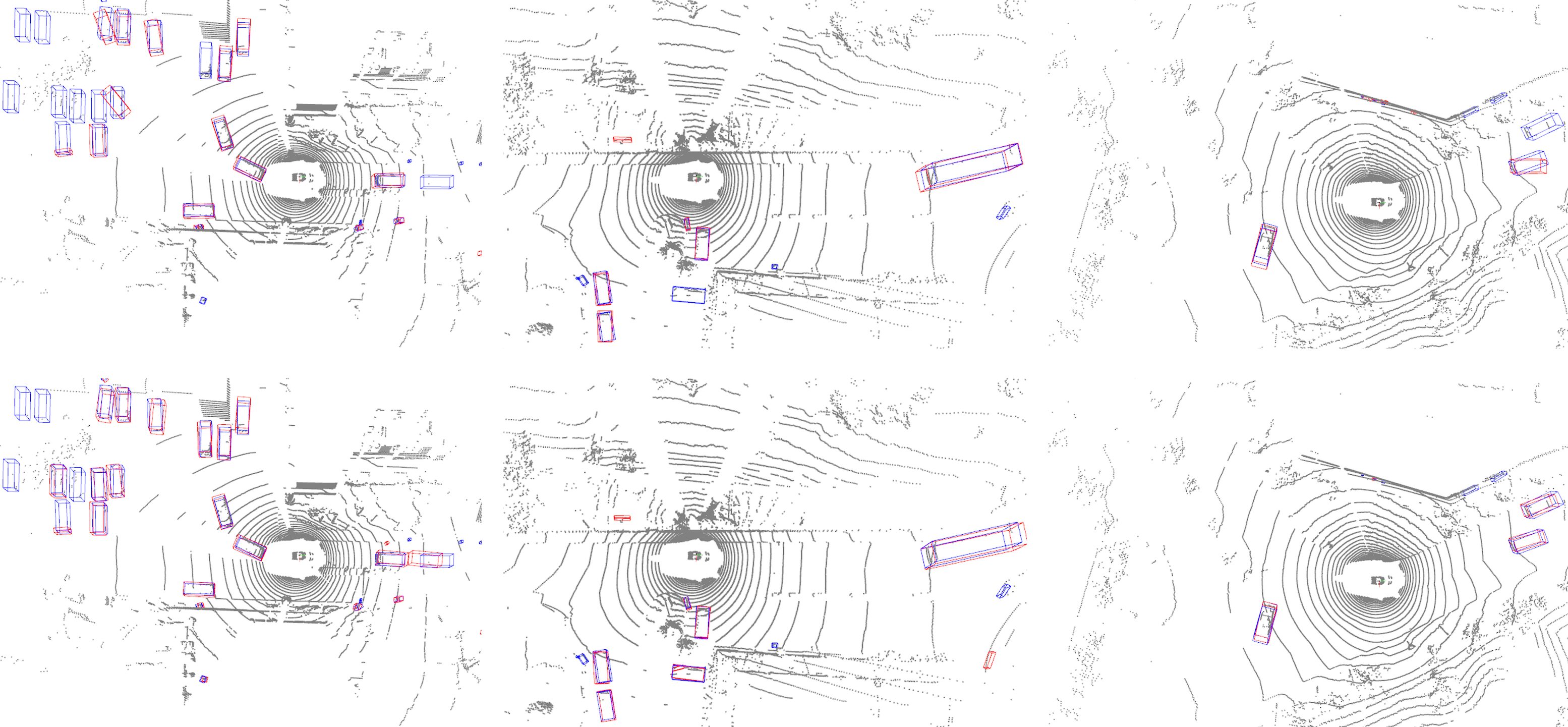}
    \end{center}
    \vspace{-0.2cm}
    \caption{\textbf{Qualitative results} of object detection trained with $5\%$ labeled data. The first row shows the model trained with random initialization, while the second row displays results from our framework. The \textcolor{blue}{groundtruth} / \textcolor{red}{predicted} results are highlighted with \textcolor{blue}{blue} / \textcolor{red}{red} boxes, respectively. Best viewed in colors and zoomed-in for additional details.}
\label{fig:qualitative_det}
\end{figure*}

\subsection{Implementation Details}
\label{sec:exp_implementation}
We utilize MinkUNet \cite{choy2019minkowski} as our 3D backbone, processing cylindrical voxels with a resolution of $0.10$m. For the 2D backbone, we adopt models from ResNet-50 \cite{he2016residual} and the ViT family \cite{dosovitskiy2020vit}, pretrained using the DINOv2 framework \cite{oquab2023dinov2}, as proposed in \cite{puy2024scalr}. The pretraining of our LiDAR point cloud encoder spans 50 epochs, using eight GPUs with a batch size of 32. We employ AdamW for optimization, coupled with a OneCycle learning rate scheduler. During fine-tuning, we strictly adhere to the data splits, augmentation strategies, and evaluation protocols from SLidR \cite{sautier2022slidr} for the \emph{nuScenes} and \emph{SemanticKITTI} datasets, extending similar procedures to other datasets. The training objective combines cross-entropy loss and Lov{\'a}sz-Softmax loss \cite{berman2018lovasz} to optimize the model. For multi-source data pretraining, we combine data from three major datasets: \emph{nuScenes}, \emph{SemanticKITTI}, and \emph{Waymo Open}. We also study pretraining with each individual dataset and their combination. Image-LiDAR pairs from these datasets are randomly sampled to form the training batches. The VFM-based superpixels (\emph{e.g.}, from SAM \cite{kirillov2023sam} and SEEM \cite{zou2023seem}) are generated offline before pretraining, and do not incur overhead during downstream training or inference. We report results of prior works based on their official publications \cite{sautier2022slidr,mahmoud2023st}. Additionally, for methods like PPKT \cite{liu2021ppkt} and SLidR \cite{sautier2022slidr}, which were only evaluated on \emph{nuScenes} and \emph{SemanticKITTI}, we reproduce their performance on the other nine datasets using publicly available code.

\noindent\textbf{Evaluation Metrics}. 
For linear probing and downstream tasks on LiDAR semantic segmentation, we report the Intersection-over-Union (IoU) for each semantic class and the mean IoU (mIoU) across all classes. For 3D object detection, we report the mean Average Precision (mAP) and nuScenes Detection Score (NDS). For panoptic LiDAR segmentation, we follow the Panoptic-nuScenes protocol~\cite{fong2022panoptic-nuScenes} and report the Panoptic Quality (PQ), Recognition Quality (RQ), and Segmentation Quality (SQ). For robustness evaluation, we adopt the Robo3D benchmark \cite{kong2023robo3D} and report the mean Corruption Error (mCE) and mean Resilience Rate (mRR) scores. The baseline model used for these evaluations is the MinkUNet implementation in \cite{tang2020searching}.

\subsection{Comparative Study}
\label{sec:comparative}

\noindent\textbf{Comparisons to State-of-the-Arts}.
We compare the proposed \textbf{LargeAD} against random initialization and a total of eleven state-of-the-art pretraining techniques using both linear probing (LP) and few-shot fine-tuning protocols on \emph{nuScenes} \cite{fong2022panoptic-nuScenes}, as presented in Table~\ref{tab:pre_nuscenes}. The results demonstrate the significant impact of pretraining on downstream task performance, particularly in low-data regimes such as $1\%$, $5\%$, and $10\%$ fine-tuning budgets. When distilling the knowledge from ResNet, ViT-S, ViT-B, and ViT-L, our framework achieves mIoU scores of $46.13\%$, $46.58\%$, $47.84\%$, and $48.71\%$ under the LP setting, outperforming the previous best models \cite{mahmoud2023st,liu2023seal,chen2024csc,puy2024scalr} by large margins. Moreover, our framework consistently delivers the highest performance across almost all fine-tuning tasks on \emph{nuScenes}, highlighting the effectiveness of VFM-assisted contrastive learning, the incorporation of spatial-temporal consistency regularization, and the combination of multi-source data pretraining.

\noindent\textbf{Downstream Generalization}.
To thoroughly evaluate the generalization capability of our method, we perform experiments across a total of nine autonomous driving datasets, with the results summarized in Table~\ref{tab:pre_nuscenes} (for \emph{SemanticKITTI} and \emph{Waymo Open}) and Table~\ref{tab:multiple_datasets} (for the other seven datasets). Each dataset presents distinct challenges, including variations in sensor types, acquisition environments, scale, and data fidelity, making this a rigorous assessment of model generalization. Our framework achieves $51.68\%$ and $52.68\%$ mIoU scores on \emph{SemanticKITTI} and \emph{Waymo Open} when distilling from ViT-L, setting new states for these benchmarks. We also surpass SLidR \cite{sautier2022slidr} and Seal \cite{liu2023seal} on the other seven datasets, as shown in Table~\ref{tab:multiple_datasets}. The results consistently show that the proposed method outperforms existing state-of-the-art approaches on all evaluated datasets. These outcomes underline the robustness and adaptability of our proposed approach to a wide range of real-world automotive perception tasks.

\input{tables/detection}
\noindent\textbf{Robustness Probing}.
Assessing the robustness of learned representations on out-of-distribution data is crucial, particularly for real-world applications where environments are unpredictable. We utilize the \emph{nuScenes-C} dataset from the Robo3D benchmark \cite{kong2023robo3D} to evaluate robustness under various corruptions. As shown in Table~\ref{tab:robustness}, self-supervised learning methods like PPKT \cite{liu2021ppkt} and SLidR \cite{sautier2022slidr} generally demonstrate better resilience compared to traditional baselines (with the random initialization) such as MinkUNet \cite{choy2019minkowski}. Our approach, \textbf{LargeAD}, achieves superior robustness across nearly all corruption types, outperforming other recent segmentation backbones that rely on different LiDAR representations, including range view \cite{cheng2022cenet}, bird’s eye view (BEV) \cite{zhou2020polarNet}, raw point-based methods \cite{puy23waffleiron}, and multi-view fusion \cite{tang2020searching}. These results underscore the adaptability and resilience of our pretraining framework under diverse real-world autonomous driving conditions.

\noindent\textbf{Enhancements for 3D Object Detection}.
In addition to LiDAR semantic segmentation, we extend our framework to 3D object detection tasks on the \emph{nuScenes} dataset \cite{caesar2020nuScenes} and compare it with state-of-the-art pretraining methods. The results, shown in Table~\ref{tab:detection}, indicate that \textbf{LargeAD} consistently outperforms competing approaches across various data proportions ($5\%$, $10\%$, and $20\%$) for both the CenterPoint \cite{centerpoint} and SECOND \cite{second} backbones. In particular, our method achieves the highest mAP and NDS at all fine-tuning levels, surpassing recent techniques such as CSC \cite{chen2024csc} and TriCC \cite{pang2023tricc}. Notably, our framework maintains superior performance with limited fine-tuning data, demonstrating its robustness and effectiveness for 3D object detection. These results further validate the generalizability of our framework across multiple challenging tasks in autonomous driving, from semantic segmentation to object detection.

\input{tables/panoptic}
\noindent\textbf{Enhancements for Panoptic LiDAR Segmentation}.
To further showcase the general-purpose capability of \textbf{LargeAD}, we evaluate its performance on the panoptic LiDAR segmentation task using the Panoptic-nuScenes benchmark \cite{fong2022panoptic-nuScenes}. This task jointly requires semantic-level understanding and instance-level separation, making it a strong indicator of representation richness. As shown in Table~\ref{tab:panoptic}, our model achieves the best results across all metrics (PQ, RQ, and SQ) under both 1\% and 5\% fine-tuning regimes. Compared to previous methods like SLidR~\cite{sautier2022slidr} and Seal~\cite{liu2023seal}, our framework demonstrates consistent advantages, particularly in instance-level recognition (RQ). These results confirm that the representations learned can effectively support both category discrimination and fine-grained object delineation, further validating our general applicability.

\noindent\textbf{Qualitative Assessment}.
To further assess the performance of our framework, we visualize the segmentation predictions on \emph{nuScenes} \cite{fong2022panoptic-nuScenes} in Figure~\ref{fig:qualitative}. The pretraining methods clearly enhance segmentation quality compared to models trained from random initialization. Among the compared approaches, \textbf{LargeAD} demonstrates the most consistent and accurate results, particularly in complex driving environments. This improvement can be attributed to the robust spatial and temporal consistency learning embedded in our pretraining strategy. Additionally, we perform qualitative comparisons of our framework with the random initialization on the 3D object detection. Qualitative results from Figure~\ref{fig:qualitative_det} further highlight our superiority. Our framework generates significantly more accurate predictions, closely matching ground truth boxes and outperforming models trained from random initialization, which exhibit greater inaccuracies. The strong performance across both tasks underscores the generalizability and robustness of our pretraining approach.

\input{tables/zero_shot_lidarseg}
\noindent\textbf{Extension to Zero-Shot LiDAR Segmentation}.
To further validate the generalization ability of our pretraining framework, we conduct zero-shot LiDAR semantic segmentation experiments following the protocol of OpenScene~\cite{peng2023openscene}. This task bypasses any label supervision by leveraging CLIP-based image-to-point alignment, where semantic predictions are made via cosine similarity with textual category prompts. As shown in Table~\ref{tab:zero-shot}, LargeAD achieves superior performance over existing methods, obtaining $43.8\%$ mIoU and $62.7\%$ mAcc on the nuScenes validation set. These results not only highlight the rich cross-modal semantics captured by our pretraining pipeline, but also reinforce its strong transferability in label-scarce or fully unsupervised deployment scenarios.

\subsection{Ablation Study}
\label{sec:ablation}

\noindent\textbf{Comparing Different Foundation Models}. 
This study serves as the first attempt of adapting VFMs for large-scale point cloud representation learning. We conduct a comprehensive ablation study on four popular VFMs, \emph{i.e.}, SAM \cite{kirillov2023sam}, X-Decoder \cite{zou2023xcoder}, OpenSeeD \cite{zhang2023openSeeD}, and SEEM \cite{zou2023seem}, and show the results in Table~\ref{tab:foundation_models}. Our experiments reveal that different VFMs have varying levels of impact on contrastive learning objectives. All VFMs consistently outperform traditional SLIC \cite{achanta2012slic} when integrated into both frameworks. SEEM \cite{zou2023seem} emerges as the top performer, providing the most consistent improvements. We attribute this to its ability to produce compact, semantically aligned superpixels that better match LiDAR segments. Interestingly, SAM \cite{kirillov2023sam} generates more fine-grained superpixels, which enhances performance when fine-tuning with larger labeled datasets. However, its finer segmentation may lead to redundant or overly fragmented contrastive pairs, explaining the lower performance under limited supervision. We hypothesize that SAM \cite{kirillov2023sam} offers more diverse negative samples, which may benefit the superpixel-driven contrastive learning. Across all configurations, our framework surpasses SLidR \cite{sautier2022slidr} by a significant margin, affirming the effectiveness of our proposed large-scale cross-sensor data pretraining framework.

\input{tables/foundation_models}
\input{tables/ablation}
\input{tables/sensors}

\noindent\textbf{Cosine Similarity}. 
We visualize feature similarities across various VFMs in Figure~\ref{fig:cosine}, providing insights into the distinction in representations even before fine-tuning. Semantically enriched models like X-Decoder \cite{zou2023xcoder}, OpenSeeD \cite{zhang2023openSeeD}, and SEEM \cite{zou2023seem} exhibit clear feature distinctions between objects and backgrounds. In contrast, unsupervised or overly fine-grained methods such as SLIC \cite{achanta2012slic} and SAM \cite{kirillov2023sam} show weaker semantic awareness. These qualitative observations are mirrored in the performance results from both linear probing and fine-tuning tasks (see Table~\ref{tab:foundation_models}), where SEEM demonstrates stronger consistency regularization during cross-sensor representation learning, leading to improved downstream task performance.

\noindent\textbf{Component Analysis}.
The ablation results of our core components are in Table~\ref{tab:ablation}. Integrating VFMs alone (row \textcolor{red}{c}) delivers a $4.20\%$ mIoU improvement in linear probing, while adding temporal consistency learning (row \textcolor{red}{b}) yields an additional $1.65\%$ mIoU gain. Combining these two components (row \textcolor{red}{d}) results in a total of $5.21\%$ mIoU boost. The point-to-segment regularization (row \textcolor{red}{e}) also contributes a significant improvement of $4.55\%$ mIoU on its own. Moreover, when both STC and P2S are used jointly (row \textcolor{red}{f}), we observe additional gains beyond their individual contributions, suggesting that the two modules offer complementary benefits. When all components are integrated (row \textcolor{red}{g}), the final model achieves a total $6.33\%$ mIoU gain over SLidR \cite{sautier2022slidr}, outperforming all state-of-the-art methods on both in-distribution and out-of-distribution benchmarks.

\input{tables/data_sources}
\noindent\textbf{Scaling with Data Sources}.
We conduct an ablation study to examine the impact of using different datasets during pretraining, as summarized in Table~\ref{tab:sources}. The results demonstrate that pretraining on a single dataset, \emph{i.e.}, \emph{nuScenes} (N) \cite{fong2022panoptic-nuScenes}, \emph{SemanticKITTI} (K) \cite{behley2019semanticKITTI}, or \emph{Waymo Open} (W) \cite{sun2020waymoOpen}, provides notable improvements over random initialization, especially in the linear probing (LP) and $1\%$ fine-tuning evaluations. However, as more datasets are combined during pretraining, the performance continues to improve consistently across both in-domain (the pretrained dataset) and out-of-domain datasets. For example, pretraining on all three datasets (N + K + W) results in the best performance, achieving a significant boost across all scenarios. Interestingly, the benefits of multi-dataset pretraining are most evident in the out-of-domain results, where combining two or three datasets leads to substantial gains over single-dataset pretraining. For instance, the inclusion of both \emph{nuScenes} and \emph{Waymo Open} (N + W) leads to a $47.42\%$ mIoU in LP on \emph{nuScenes}, outperforming the single-dataset pretraining setups. Similarly, using all three datasets outperforms two-dataset combinations in both LP and fine-tuning, particularly in out-of-domain scenarios like \emph{Waymo Open}, where it achieves a remarkable $51.52\%$ mIoU in $1\%$ fine-tuning. These results highlight the importance of multi-source pretraining, which not only improves generalization within in-domain datasets but also significantly enhances out-of-domain performance, showcasing the robustness and scalability of our proposed framework.

\begin{wrapfigure}{r}{0.5\textwidth}
    \begin{minipage}{\linewidth}
        \centering
        \vspace{-0.55cm}
        \includegraphics[width=\linewidth]{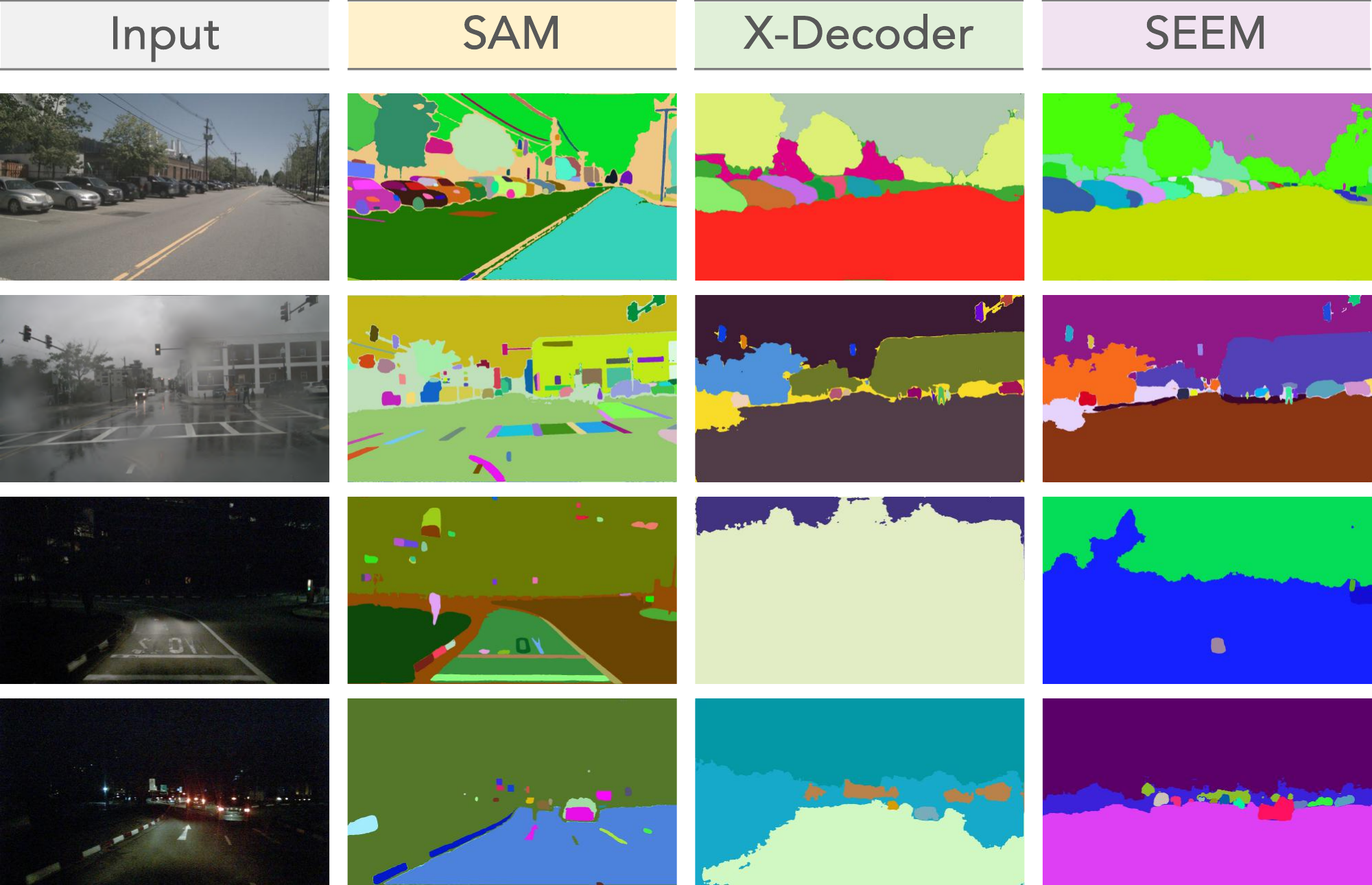}
        \vspace{-0.55cm}
        \caption{\textbf{Failure cases of superpixels} generated by different VFMs, including SAM \cite{kirillov2023sam}, SEEM \cite{zou2023seem}, and X-Decoder \cite{zou2023xcoder}, under various challenging conditions such as occlusion, adverse weather, and nighttime scenes. Different colors in an image represent different segmentation instances.}
        \label{fig:vfm_failure}
        \vspace{0.1cm}
    \end{minipage}
\end{wrapfigure}
\noindent\textbf{Robustness to Sensor Misalignment}. 
To evaluate the robustness of our framework to calibration errors between LiDAR and cameras, we simulate misalignment by perturbing the extrinsic transformation matrix $\Gamma_K$ with random translation and rotation noise at three levels: 1\%, 5\%, and 10\%. As in Table~\ref{tab:sensors}, our approach exhibits strong resilience under misalignment, with minimal performance degradation even at 10\% perturbation. This robustness can be attributed to the region-level supervision provided by VFM–generated superpixels, which are less sensitive to precise point-to-pixel alignment, and to the superpoint temporal consistency module, which operates entirely in 3D space. In contrast, point-wise methods such as PPKT~\cite{liu2021ppkt} suffer larger performance drops due to their reliance on accurate 2D–3D correspondences. These results suggest that our framework remains effective even under mild to moderate calibration noise, enhancing its applicability in real-world deployments.

\begin{wrapfigure}{r}{0.5\textwidth}
    \begin{minipage}{\linewidth}
        \centering
        \vspace{-0.9cm}
        \includegraphics[width=\linewidth]{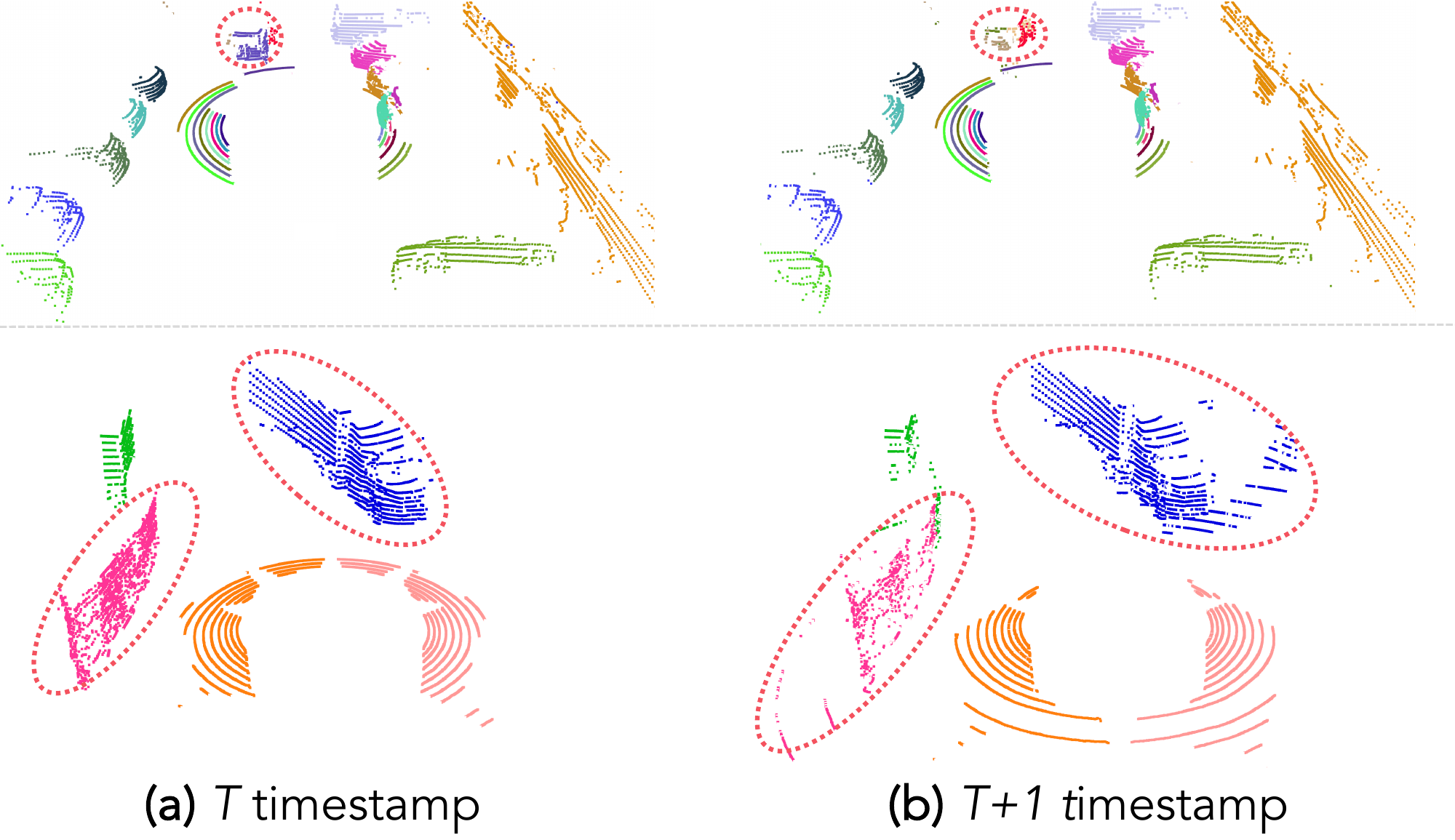}
        \vspace{-0.55cm}
        \caption{\textbf{Temporal failure cases} segmented by RANSAC \cite{foschler1981ransac}. In the \textbf{first row}, dynamic objects are segmented into inconsistent instances across different timestamps, indicating a lack of temporal coherence. In the \textbf{second row}, the moving objects are either incorrectly merged with surrounding background points or fragmented into multiple inconsistent instances due to motion and scene clutter.}
        \label{fig:temporal_failure}
        \vspace{-1.1cm}
    \end{minipage}
\end{wrapfigure}
\noindent\textbf{Visual Examples of Failure Cases}.
In addition to the above discussions, we have included the corresponding visual examples in Figure~\ref{fig:vfm_failure} and Figure~\ref{fig:temporal_failure} for further reference. Specifically, we show superpixel failures from different VFMs under occlusion, nighttime, and adverse weather scenes, as well as inconsistent temporal segmentation across frames. These examples offer additional insight into the failure modes of the proposed modules and complement previous discussions.

%% file: tables/pre_semantickitti.tex
\begin{table*}[t]
    \centering
    \caption{Benchmarking data pretraining methods that are: (a) random initialized, (b) pretrained on \emph{SemanticKITTI} \cite{behley2019semanticKITTI}, (c) pretrained on \emph{Waymo} \cite{sun2020waymoOpen}, and (d) pretrained based on our large-scale cross-sensor data pretraining pipeline. After pretraining, all methods are fine-tuned on \emph{SemanticKITTI} \cite{behley2019semanticKITTI}, \emph{nuScenes} \cite{caesar2020nuScenes}, and \emph{Waymo} \cite{sun2020waymoOpen}, respectively, and evaluated on the official validation split of each dataset. \textbf{LP} denotes linear probing with frozen backbones. All mIoU scores are given in percentage (\%). The \textbf{best} and \underline{second best} scores are highlighted in \textbf{bold} and \underline{underline}.}
    \vspace{-0.1cm}
\label{tab:pre_semantickitti}
\resizebox{\linewidth}{!}{
    \begin{tabular}{c|r|r|c|cccc|cccc|cccc}
    \toprule
    \multirow{2}{*}{\textbf{\#}} & \multirow{2}{*}{\textbf{Method}} & \multirow{2}{*}{\textbf{Venue}} & \multirow{2}{*}{\textbf{Distill}} & \multicolumn{4}{c}{\textbf{SemanticKITTI}} \vline & \multicolumn{4}{c}{\textbf{nuScenes}} \vline & \multicolumn{4}{c}{\textbf{Waymo}}
    \\
    & & & & \textbf{LP} & \textbf{1\%} & \textbf{10\%} & \textbf{Full} & \textbf{LP} & \textbf{1\%} & \textbf{10\%} & \textbf{Full} & \textbf{LP} & \textbf{1\%} & \textbf{10\%} & \textbf{Full}
    \\\midrule\midrule
    (a) & \cellcolor{yellow!8}Random & \cellcolor{yellow!8}- & \cellcolor{yellow!8}- & \cellcolor{yellow!8}7.45 & \cellcolor{yellow!8}39.50 & \cellcolor{yellow!8}55.11 & \cellcolor{yellow!8}57.39 & \cellcolor{yellow!8}8.10 & \cellcolor{yellow!8}30.30 & \cellcolor{yellow!8}56.15 & \cellcolor{yellow!8}74.66 & \cellcolor{yellow!8}6.45 & \cellcolor{yellow!8}39.41 & \cellcolor{yellow!8}59.71 & \cellcolor{yellow!8}64.22
    \\\midrule
    \multirow{2}{*}{(b)} & SLidR \cite{sautier2022slidr} & CVPR'22 & ViT-B & 32.83 & 45.20 & 56.18 & 56.60 & 16.54 & 35.21 & 56.77 & 74.47 & 27.82 & 44.35 & 60.55 & 64.33
    \\
    & Seal \cite{liu2023seal} & NeurIPS'23 & ViT-B & \underline{33.46} & \underline{47.52} & \underline{57.68} & 58.64 & 18.18 & 38.86 & 58.05 & \underline{74.53} & 29.36 & 45.96 & 61.59 & 64.81
    \\\midrule
    \multirow{2}{*}{(c)} & SLidR \cite{sautier2022slidr} & CVPR'22 & ViT-B & 32.29 & 44.75 & 55.66 & 57.49 & 23.97 & 39.16 & 58.47 & 73.96 & 32.21 & 47.96 & 61.33 & 64.43
    \\
    & Seal \cite{liu2023seal} & NeurIPS'23 & ViT-B & 33.21 & 47.46 & 57.61 & \underline{58.72} & \underline{25.74} & \underline{42.69} & \underline{60.59} & 74.38 & \underline{34.65} & \underline{49.30} & \underline{62.37} & \underline{65.79}
    \\\midrule
    (d) & \cellcolor{violet!7}\textcolor{violet!66}{\textbf{LargeAD}} & \cellcolor{violet!7}\textbf{Ours} & \cellcolor{violet!7}ViT-B & \cellcolor{violet!7}\textbf{35.77} & \cellcolor{violet!7}\textbf{50.68} & \cellcolor{violet!7}\textbf{60.18} & \cellcolor{violet!7}\textbf{60.98} & \cellcolor{violet!7}\textbf{47.84} & \cellcolor{violet!7}\textbf{48.37} & \cellcolor{violet!7}\textbf{64.11} & \cellcolor{violet!7}\textbf{75.91} & \cellcolor{violet!7}\textbf{36.12} & \cellcolor{violet!7}\textbf{51.52} & \cellcolor{violet!7}\textbf{63.92} & \cellcolor{violet!7}\textbf{66.10}
    \\
    \bottomrule
\end{tabular}
}
\end{table*}

%% file: tables/multiple_datasets.tex
\begin{table*}[t]
    \centering
    \caption{Comparisons of different data pretraining methods that are: (a) random initialized, (b) pretrained on \emph{nuScenes} \cite{fong2022panoptic-nuScenes}, (c) pretrained on \emph{SemanticKITTI} \cite{behley2019semanticKITTI}, (d) pretrained on \emph{Waymo Open} \cite{sun2020waymoOpen}, and (e) pretrained based on the proposed large-scale cross-sensor data pretraining pipeline. After pretraining, all methods are fine-tuned on different downstream datasets \cite{unal2022scribbleKITTI,jiang2021rellis3D,pan2020semanticPOSS,xiao2023semanticSTF,xiao2022synLiDAR,klokov2023daps3D,saltori2020synth4D}, respectively, and evaluated on the official validation split of each dataset. All mIoU scores are given in percentage (\%). The \textbf{best} and \underline{second best} scores are highlighted in \textbf{bold} and \underline{underline}.}
    \vspace{-0.1cm}
\label{tab:multiple_datasets}
\resizebox{\linewidth}{!}{
    \begin{tabular}{c|r|cc|cc|cc|cc|cc|cc|cc}
    \toprule
    \multirow{2}{*}{\textbf{\#}} & \multirow{2}{*}{\textbf{Method}} & \multicolumn{2}{c}{\textbf{ScribbleKITTI}} \vline & \multicolumn{2}{c}{\textbf{RELLIS-3D}} \vline & \multicolumn{2}{c}{\textbf{SemanticPOSS}} \vline & \multicolumn{2}{c}{\textbf{SemanticSTF}} \vline & \multicolumn{2}{c}{\textbf{SynLiDAR}} \vline & \multicolumn{2}{c}{\textbf{DAPS-3D}} \vline & \multicolumn{2}{c}{\textbf{Synth4D}}
    \\
    & & {\textbf{1\%}} & {\textbf{10\%}} & {\textbf{1\%}} & {\textbf{10\%}} & {\textbf{Half}} & {\textbf{Full}} & {\textbf{Half}} & {\textbf{Full}} & {\textbf{1\%}} & {\textbf{10\%}} & {\textbf{Half}} & {\textbf{Full}} & {\textbf{1\%}} & {\textbf{10\%}}
    \\\midrule\midrule
    (a) & \cellcolor{yellow!8}Random & \cellcolor{yellow!8}23.81 & \cellcolor{yellow!8}47.60 & \cellcolor{yellow!8}38.46 & \cellcolor{yellow!8}53.60 & \cellcolor{yellow!8}46.26 & \cellcolor{yellow!8}54.12 & \cellcolor{yellow!8}48.03 & \cellcolor{yellow!8}48.15 & \cellcolor{yellow!8}19.89 & \cellcolor{yellow!8}44.74 & \cellcolor{yellow!8}74.32 & \cellcolor{yellow!8}79.38 & \cellcolor{yellow!8}20.22 & \cellcolor{yellow!8}66.87
    \\\midrule
    \multirow{3}{*}{(b)} & PPKT \cite{liu2021ppkt} & 36.50 & 51.67 & 49.71 & 54.33 & 50.18 & 56.00 & 50.92 & 54.69 & 37.57 & 46.48 & 78.90 & 84.00 & 61.10 & 62.41
    \\
    & SLidR \cite{sautier2022slidr} & 39.60 & 50.45 & 49.75 & 54.57 & 51.56 & 55.36 & 52.01 & 54.35 & 42.05 & 47.84 & 81.00 & 85.40 & 63.10 & 62.67
    \\
    & Seal \cite{liu2023seal} & 40.64 & 52.77 & 51.09 & 55.03 & 53.26 & 56.89 & 53.46 & 55.36 & 43.58 & 49.26 & 81.88 & 85.90 & 64.50 & 66.96
    \\\midrule
    \multirow{2}{*}{(c)} & SLidR \cite{sautier2022slidr} & 40.27 & 51.92 & 47.65 & 54.03 & 49.97 & 54.83 & 51.39 & 53.83 & 40.11 & 43.90 & 74.93 & 80.31 & 57.24 & 61.25 
    \\
    & Seal \cite{liu2023seal} & 41.97 & 53.25 & 49.23 & 54.93 & 52.31 & 55.47 & 53.26 & 55.49 & 41.48 & 45.52 & 76.31 & 81.24 & 59.25 & 64.21 
    \\\midrule
    \multirow{2}{*}{(d)} & SLidR \cite{sautier2022slidr} & 38.24 & 48.96 & 47.31 & 53.20 & 50.93 & 55.04 & 50.93 & 54.35 & 43.91 & 49.32 & 79.02 & 84.34 & 61.98 & 62.15 
    \\
    & Seal \cite{liu2023seal} & 39.13 & 50.53 & 50.55 & 54.68 & 52.90 & 55.27 & 52.20 & 55.24 & 44.78 & 50.98 & 79.68 & 84.68 & 63.74 & 65.39
    \\\midrule
    (e) & \cellcolor{violet!7}\textcolor{violet!66}{\textbf{LargeAD}} & \cellcolor{violet!7}\textbf{43.45} & \cellcolor{violet!7}\textbf{54.62} & \cellcolor{violet!7}\textbf{53.31} & \cellcolor{violet!7}\textbf{56.62} & \cellcolor{violet!7}\textbf{54.47} & \cellcolor{violet!7}\textbf{57.11} & \cellcolor{violet!7}\textbf{55.14} & \cellcolor{violet!7}\textbf{56.89} & \cellcolor{violet!7}\textbf{47.12} & \cellcolor{violet!7}\textbf{52.82} & \cellcolor{violet!7}\textbf{83.31} & \cellcolor{violet!7}\textbf{86.21} & \cellcolor{violet!7}\textbf{65.33} & \cellcolor{violet!7}\textbf{67.12}
    \\
    \bottomrule
\end{tabular}
}
\end{table*}

%% file: tables/robustness.tex
\begin{table*}[t]
    \centering
    \caption{Robustness probing of data pretraining methods under eight out-of-distribution corruptions in the \emph{nuScenes-C} dataset from the Robo3D benchmark \cite{kong2023robo3D}. The \textbf{mCE} score is the lower the better while \textbf{mRR} and \textbf{mIoU} scores are the higher the better. All mCE, mRR, and mIoU scores are given in percentage (\%). \textbf{Avg} denotes the average mIoU scores of methods across all eight corruptions. The \textbf{best} and \underline{second best} scores are highlighted in \textbf{bold} and \underline{underline}.}
    \vspace{-0.1cm}
\label{tab:robustness}
\resizebox{\linewidth}{!}{
    \begin{tabular}{c|r|c|p{0.8cm}<{\centering}p{0.8cm}<{\centering}|p{0.8cm}<{\centering}p{0.8cm}<{\centering}p{0.8cm}<{\centering}p{0.8cm}<{\centering}p{0.8cm}<{\centering}p{0.8cm}<{\centering}p{0.8cm}<{\centering}p{0.8cm}<{\centering}|p{0.8cm}<{\centering}}
    \toprule
    \textbf{\#} & \textbf{Initial} & \textbf{Backbone} & \textbf{mCE} & \textbf{mRR} & \textbf{Fog} & \textbf{Wet} & \textbf{Snow} & \textbf{Move} & \textbf{Beam} & \textbf{Cross} & \textbf{Echo} & \textbf{Sensor} & \textbf{Avg}
    \\\midrule\midrule
    \multirow{4}{*}{\rotatebox[origin=c]{90}{\textbf{LP}}}
    & PPKT \cite{liu2021ppkt} & MinkU-34 & 183.44 & \textbf{78.15} & 30.65 & 35.42 & 28.12 & 29.21 & 32.82 & 19.52 & 28.01 & 20.71 & 28.06
    \\
    & SLidR \cite{sautier2022slidr} & MinkU-34 & 179.38 & 77.18 & 34.88 & 38.09 & \underline{32.64} & 26.44 & 33.73 & \underline{20.81} & 31.54 & 21.44 & 29.95
    \\
    & Seal \cite{liu2023seal} & MinkU-34 & \underline{166.18} & 75.38 & \underline{37.33} & \underline{42.77} & 29.93 & \underline{37.73} & \textbf{40.32} & 20.31 & \underline{37.73} & \underline{24.94} & \underline{33.88}
    \\
    & \cellcolor{violet!7}\textcolor{violet!66}{\textbf{LargeAD}} & \cellcolor{violet!7}MinkU-34 & \cellcolor{violet!7}\textbf{160.28} & \cellcolor{violet!7}\underline{77.29} & \cellcolor{violet!7}\textbf{40.54} & \cellcolor{violet!7}\textbf{43.26} & \cellcolor{violet!7}\textbf{37.92} & \cellcolor{violet!7}\textbf{38.27} & \cellcolor{violet!7}\underline{40.27} & \cellcolor{violet!7}\textbf{25.67} & \cellcolor{violet!7}\textbf{39.26} & \cellcolor{violet!7}\textbf{30.62} & \cellcolor{violet!7}\textbf{36.98}
    \\\midrule
    \multirow{10}{*}{\rotatebox[origin=c]{90}{\textbf{Full}}} & Random & PolarNet & 115.09 & 76.34 & 58.23 & 69.91 & 64.82 & 44.60 & 61.91 & 40.77 & 53.64 & 42.01 & 54.49
    \\
    & Random & CENet & 112.79 & 76.04 & 67.01 & 69.87 & 61.64 & 58.31 & 49.97 & \textbf{60.89} & 53.31 & 24.78 & 55.72
    \\
    & Random & WaffleIron & 106.73 & 72.78 & 56.07 & \underline{73.93} & 49.59 & 59.46 & 65.19 & 33.12 & \underline{61.51} & \underline{44.01} & 55.36
    \\
    & Random & Cylinder3D & 105.56 & 78.08 & 61.42 & 71.02 & 58.40 & 56.02 & 64.15 & 45.36 & 59.97 & 43.03 & 57.42
    \\
    & Random & SPVCNN-34 & 106.65 & 74.70 & 59.01 & 72.46 & 41.08 & 58.36 & 65.36 & 36.83 & \textbf{62.29} & \textbf{49.21} & 55.58
    \\
    & \cellcolor{yellow!8}Random & \cellcolor{yellow!8}MinkU-34 & \cellcolor{yellow!8}112.20 & \cellcolor{yellow!8}72.57 & \cellcolor{yellow!8}62.96 & \cellcolor{yellow!8}70.65 & \cellcolor{yellow!8}55.48 & \cellcolor{yellow!8}51.71 & \cellcolor{yellow!8}62.01 & \cellcolor{yellow!8}31.56 & \cellcolor{yellow!8}59.64 & \cellcolor{yellow!8}39.41 & \cellcolor{yellow!8}54.18
    \\
    & PPKT \cite{liu2021ppkt} & MinkU-34 & 105.64 & 76.06 & 64.01 & 72.18 & 59.08 & 57.17 & 63.88 & 36.34 & 60.59 & 39.57 & 56.60
    \\
    & SLidR \cite{sautier2022slidr} & MinkU-34 & 106.08 & 75.99 & 65.41 & 72.31 & 56.01 & 56.07 & 62.87 & 41.94 & 61.16 & 38.90 & 56.83
    \\
    & Seal \cite{liu2023seal} & MinkU-34 & \underline{92.63} & \underline{83.08} & \textbf{72.66} & \textbf{74.31} & \underline{66.22} & \textbf{66.14} & \underline{65.96} & 57.44 & 59.87 & 39.85 & \underline{62.81}
    \\
    & \cellcolor{violet!7}\textcolor{violet!66}{\textbf{LargeAD}} & \cellcolor{violet!7}MinkU-34 & \cellcolor{violet!7}\textbf{91.75} & \cellcolor{violet!7}\textbf{83.61} & \cellcolor{violet!7}\underline{71.95} & \cellcolor{violet!7}72.47 & \cellcolor{violet!7}\textbf{67.28} & \cellcolor{violet!7}\underline{65.29} & \cellcolor{violet!7}\textbf{67.49} & \cellcolor{violet!7}\underline{59.42} & \cellcolor{violet!7}61.38 & \cellcolor{violet!7}42.46 & \cellcolor{violet!7}\textbf{63.47}
    \\
    \bottomrule
\end{tabular}
}
\end{table*}

%% file: tables/detection.tex
\begin{wraptable}{r}{0.5\textwidth}
\centering
\vspace{-0.1cm}
    \centering
    \caption{Comparisons of state-of-the-art pretraining methods pretrained and fine-tuned on \emph{nuScenes} \cite{caesar2020nuScenes} using the specified data proportions. All methods employ CenterPoint \cite{centerpoint} or SECOND \cite{second} as the 3D object detection backbones. All sores are given in percentage (\%). The \textbf{best} and \underline{second best} scores are highlighted in \textbf{bold} and \underline{underline}.}
    \vspace{-0.1cm}
\label{tab:detection}
\resizebox{\linewidth}{!}{
\begin{tabular}{r|p{0.6cm}<{\centering}p{0.6cm}<{\centering}|p{0.6cm}<{\centering}p{0.6cm}<{\centering}|p{0.6cm}<{\centering}p{0.6cm}<{\centering}}
    \toprule
    \multirow{3.7}{*}{\textbf{Method}} & \multicolumn{6}{c}{\textbf{nuScenes}}
    \\\cmidrule{2-7}
    & \multicolumn{2}{c|}{\textbf{5\%}} & \multicolumn{2}{c|}{\textbf{10\%}}& \multicolumn{2}{c}{\textbf{20\%}}
    \\
    & \textbf{mAP} & \textbf{NDS} & \textbf{mAP} & \textbf{NDS} & \textbf{mAP} & \textbf{NDS}
    \\\midrule\midrule
    \multicolumn{7}{c}{\emph{Backbone: VoxelNet + CenterPoint}}
    \\\midrule
    \cellcolor{yellow!8}Random & \cellcolor{yellow!8}38.0 & \cellcolor{yellow!8}44.3 & \cellcolor{yellow!8}46.9 & \cellcolor{yellow!8}55.5 & \cellcolor{yellow!8}50.2 & \cellcolor{yellow!8}59.7
    \\
    PointContrast~\cite{xie2020pointContrast} & 39.8 & 45.1 & 47.7 & 56.0 & - & -
    \\
    GCC-3D~\cite{liang2021exploring} & 41.1 & 46.8 & 48.4 & 56.7 & - & -
    \\
    SLidR~\cite{sautier2022slidr} & 43.3 & 52.4 & 47.5 & 56.8 & 50.4 & 59.9
    \\
    TriCC~\cite{pang2023tricc} & 44.6 & \underline{54.4} & 48.9 & 58.1 & 50.9 & 60.3
    \\
    Seal~\cite{liu2023seal} & 44.7 & 54.0 & 49.0 & 58.1 & 51.4 & 60.7
    \\
    CSC~\cite{chen2024csc} & \underline{45.3} & 54.2 & \underline{49.3} & \underline{58.3} & \underline{51.9} & \underline{61.3}
    \\
    \cellcolor{violet!7}\textcolor{violet!66}{\textbf{LargeAD}} & \cellcolor{violet!7}\textbf{45.8} & \cellcolor{violet!7}\textbf{54.6} & \cellcolor{violet!7}\textbf{49.7} & \cellcolor{violet!7}\textbf{58.8} & \cellcolor{violet!7}\textbf{52.1} & \cellcolor{violet!7}\textbf{61.4}
    \\\midrule\midrule
    \multicolumn{7}{c}{\emph{Backbone: VoxelNet + SECOND}}
    \\\midrule
    \cellcolor{yellow!8}Random & \cellcolor{yellow!8}35.8 & \cellcolor{yellow!8}45.9 & \cellcolor{yellow!8}39.0 & \cellcolor{yellow!8}51.2 & \cellcolor{yellow!8}43.1 & \cellcolor{yellow!8}55.7
    \\
    SLidR~\cite{sautier2022slidr} & 36.6 & 48.1 & 39.8 & 52.1 & 44.2 & 56.3
    \\
    TriCC~\cite{pang2023tricc} & 37.8 & \textbf{50.0} & 41.4 & 53.5 & 45.5 & 57.7
    \\
    Seal~\cite{liu2023seal} & 37.9 & 49.4 & 41.9 & 54.0 & 44.9 & 57.4
    \\
    CSC~\cite{chen2024csc} & \underline{38.2} & 49.4 & \underline{42.5} & \underline{54.8} & \underline{45.6} & \underline{58.1}
    \\
    \cellcolor{violet!7}\textcolor{violet!66}{\textbf{LargeAD}} & \cellcolor{violet!7}\textbf{38.3} & \cellcolor{violet!7}\underline{49.9} & \cellcolor{violet!7}\textbf{43.0} & \cellcolor{violet!7}\textbf{55.0} & \cellcolor{violet!7}\textbf{45.7} & \cellcolor{violet!7}\textbf{58.7}
    \\\bottomrule
\end{tabular}}
\vspace{-0.2cm}
\end{wraptable}

%% file: tables/panoptic.tex
\begin{wraptable}{r}{0.59\textwidth}
\centering
\vspace{-0.5cm}
    \centering
    \caption{Comparisons of state-of-the-art pretraining methods \cite{sautier2022slidr,liu2021ppkt,liu2023seal} for the panoptic LiDAR segmentation task on the \emph{Panoptic-nuScenes} \cite{fong2022panoptic-nuScenes} dataset. All scores are given in percentage (\%). The \textbf{best} and \underline{second best} scores are highlighted in \textbf{bold} and \underline{underline}.}
    \vspace{-0.1cm}
    \resizebox{\linewidth}{!}{
    \begin{tabular}{r|cccc|cccc}
    \toprule
    \multirow{2}{*}{\textbf{Method}}
    & \multicolumn{4}{c|}{\textbf{1\%}} & \multicolumn{4}{c}{\textbf{5\%}} 
    \\
    & \textbf{PQ} & \textbf{$\text{PQ}^\dagger$} & \textbf{RQ} & \textbf{SQ}
    & \textbf{PQ} & \textbf{$\text{PQ}^\dagger$} & \textbf{RQ} & \textbf{SQ}
    \\
    \midrule\midrule
    \cellcolor{yellow!8}Random & \cellcolor{yellow!8}23.10 & \cellcolor{yellow!8}28.26 & \cellcolor{yellow!8}29.78 & \cellcolor{yellow!8}58.65 & \cellcolor{yellow!8}40.36 & \cellcolor{yellow!8}44.82 & \cellcolor{yellow!8}49.54 & \cellcolor{yellow!8}78.06
    \\
    PPKT \cite{liu2021ppkt} & 30.47 & 34.75 & 36.06 & 61.19 & 41.90 & 45.88 & 51.26 & 78.80
    \\
    SLidR \cite{sautier2022slidr} & 35.10 & 39.42 & 43.65 & 71.51 & 44.04 & 47.89 & 53.57 & 79.36  
    \\
    Seal \cite{liu2023seal} & \underline{36.48} & \underline{40.31} & \underline{44.67} & \underline{72.02} & \underline{46.11} & \underline{48.94} & \underline{55.40} & \underline{80.91} 
    \\
    \cellcolor{violet!7}\textcolor{violet!66}{\textbf{LargeAD}} & \cellcolor{violet!7}\textbf{37.80} & \cellcolor{violet!7}\textbf{41.52} & \cellcolor{violet!7}\textbf{45.23} & \cellcolor{violet!7}\textbf{72.52} & \cellcolor{violet!7}\textbf{47.16} & \cellcolor{violet!7}\textbf{49.73} & \cellcolor{violet!7}\textbf{56.39} & \cellcolor{violet!7}\textbf{81.70} \\
    \bottomrule
    \end{tabular}}
\label{tab:panoptic}
\vspace{-0.2cm}
\end{wraptable}

%% file: tables/zero_shot_lidarseg.tex
\begin{wraptable}{r}{0.59\textwidth}
\centering
\vspace{-0.1cm}
    \centering
    \caption{Comparisons of zero-shot LiDAR segmentation performance on \emph{nuScenes} \cite{fong2022panoptic-nuScenes}. All mIoU and mAcc scores are given in percentage (\%). The \textbf{best} and \underline{second best} scores are highlighted in \textbf{bold} and \underline{underline}.}
    \vspace{-0.1cm}
    \label{tab:zero-shot}
    \resizebox{\linewidth}{!}{
    \begin{tabular}{r|r|cc|cc}
        \toprule
        \textbf{Method} & \textbf{Venue} & \textbf{mIoU} & \emph{Improve}~$\uparrow$ & \textbf{mAcc} & \emph{Improve}~$\uparrow$
        \\\midrule\midrule
        OpenScene \cite{peng2023openscene} & CVPR'23 & 42.1 & +0.0 & 61.8 & +0.0
        \\
        \emph{w/} Seal \cite{liu2023seal} & NeurIPS'23 & \underline{43.3} & +1.2 & \underline{62.4} & +0.6
        \\\midrule
        \cellcolor{violet!7}\emph{w/} \textcolor{violet!66}{\textbf{LargeAD}} & \cellcolor{violet!7}\textbf{Ours} & \cellcolor{violet!7}\textbf{43.8} & \cellcolor{violet!7}\textbf{+1.7} & \cellcolor{violet!7}\textbf{62.7} & \cellcolor{violet!7}\textbf{+0.9}
        \\\bottomrule
    \end{tabular}}
\label{tab:zero_shot_lidarseg}
\vspace{-0.2cm}
\end{wraptable}

%% file: tables/foundation_models.tex
\begin{table*}[t]
    \centering
    \caption{Ablation study on knowledge transfer effects from the classic SLIC \cite{achanta2012slic} algorithm and different vision foundation models \cite{kirillov2023sam,zou2023xcoder,zhang2023openSeeD,zou2023seem}. Both SLidR \cite{sautier2022slidr} and our framework are pretrained on \emph{nuScenes} \cite{fong2022panoptic-nuScenes} and fine-tuned on \emph{nuScenes} \cite{caesar2020nuScenes}, \emph{SemanticKITTI} \cite{behley2019semanticKITTI}, \emph{Waymo Open} \cite{sun2020waymoOpen}, \emph{SemanticPOSS} \cite{pan2020semanticPOSS}, \emph{SemanticSTF} \cite{bijelic2020stf}, and \emph{Synth4D} \cite{saltori2020synth4D}, respectively, and evaluated on the official validation split of each dataset. \textbf{LP} denotes linear probing with frozen backbones. All mIoU scores are given in percentage (\%). The \textbf{best} and \underline{second best} scores are highlighted in \textbf{bold} and \underline{underline}.}
    \vspace{-0.1cm}
\label{tab:foundation_models}
\resizebox{\linewidth}{!}{
    \begin{tabular}{c|c|p{0.75cm}<{\centering}p{0.75cm}<{\centering}p{0.75cm}<{\centering}p{0.75cm}<{\centering}p{0.75cm}<{\centering}p{0.75cm}<{\centering}|p{0.87cm}<{\centering}|p{0.87cm}<{\centering}|p{0.87cm}<{\centering}|p{0.87cm}<{\centering}|p{0.87cm}<{\centering}}
    \toprule
    \multirow{2}{*}{\textbf{Method}} & \multirow{2}{*}{\textbf{Superpixel}} & \multicolumn{6}{c}{\textbf{nuScenes}} \vline & \textbf{KITTI} & \textbf{Waymo} & \textbf{POSS} & \textbf{STF} & \textbf{Syn4D} 
    \\
    & & \textbf{LP} & \textbf{1\%} & \textbf{5\%} & \textbf{10\%} & \textbf{25\%} & \textbf{Full} & \textbf{1\%} & \textbf{1\%} & \textbf{Half} & \textbf{Half} & \textbf{1\%}
    \\\midrule\midrule
    \cellcolor{yellow!8}Random & \cellcolor{yellow!8}- & \cellcolor{yellow!8}8.10 & \cellcolor{yellow!8}\cellcolor{yellow!8}30.30 & \cellcolor{yellow!8}47.84 & \cellcolor{yellow!8}56.15 & \cellcolor{yellow!8}65.48 & \cellcolor{yellow!8}74.66 & \cellcolor{yellow!8}39.50 & \cellcolor{yellow!8}39.41 & \cellcolor{yellow!8}46.26 & \cellcolor{yellow!8}48.03 & \cellcolor{yellow!8}20.22
    \\\midrule
    \multirow{5}{*}{SLidR} & \cellcolor{gray!9}SLIC \cite{achanta2012slic} & \cellcolor{gray!9}38.80 & \cellcolor{gray!9}38.30 & \cellcolor{gray!9}52.49 & \cellcolor{gray!9}59.84 & \cellcolor{gray!9}66.91 & \cellcolor{gray!9}74.79 & \cellcolor{gray!9}44.60 & \cellcolor{gray!9}47.12 & \cellcolor{gray!9}51.56 & \cellcolor{gray!9}52.01 & \cellcolor{gray!9}63.10
    \\
    & SAM \cite{kirillov2023sam} & 41.49 & 43.67 & \textbf{55.97} & \textbf{61.74} & \textbf{68.85} & \textbf{75.40} & 43.35 & 48.64 & 51.37 & 52.12 & 63.15
    \\
    & X-Decoder \cite{zou2023xcoder} & 41.71 & 43.02 & \underline{54.24} & \underline{61.32} & 67.35 & 75.11 & 45.70 & \underline{48.73} & 51.50 & 52.28 & \underline{63.21}
    \\
    & OpenSeeD \cite{zhang2023openSeeD} & \underline{42.61} & \underline{43.82} & 54.17 & 61.03 & 67.30 & 74.85 & \textbf{45.88} & 48.64 & \textbf{52.09} & \textbf{52.37} & \textbf{63.31}
    \\
    & SEEM \cite{zou2023seem} & \textbf{43.00} & \textbf{44.02} & 53.03 & 60.84 & \underline{67.38} & \underline{75.21} & \underline{45.72} & \textbf{48.75} & \underline{52.00} & \underline{52.36} & 63.13
    \\\midrule
    \multirow{5}{*}{\textcolor{violet!66}{\textbf{Ours}}} & \cellcolor{gray!9}SLIC \cite{achanta2012slic} & \cellcolor{gray!9}40.89 & \cellcolor{gray!9}39.77 & \cellcolor{gray!9}53.33 & \cellcolor{gray!9}61.58 & \cellcolor{gray!9}67.78 & \cellcolor{gray!9}75.32 & \cellcolor{gray!9}45.75 & \cellcolor{gray!9}47.74 & \cellcolor{gray!9}52.77 & \cellcolor{gray!9}52.91 & \cellcolor{gray!9}63.37
    \\
    & SAM \cite{kirillov2023sam} & 43.94 & \underline{45.09} & \textbf{56.95} & 62.35 & \textbf{69.08} & \textbf{75.92} & \underline{46.53} & 49.00 & \underline{53.21} & 53.37 & 63.76
    \\
    & X-Decoder \cite{zou2023xcoder} & 42.64 & 44.31 & 55.18 & 62.03 & 68.24 & 75.56 & 46.02 & \underline{49.11} & 53.17 & \underline{53.40} & 64.21
    \\
    & OpenSeeD \cite{zhang2023openSeeD} & \underline{44.67} & 44.74 & 55.13 & \underline{62.36} & \underline{69.00} & \underline{75.64} & 46.13 & 48.98 & 53.01 & 53.27 & \underline{64.29}
    \\
    & \cellcolor{violet!7}SEEM \cite{zou2023seem} & \cellcolor{violet!7}\textbf{44.95} & \cellcolor{violet!7}\textbf{45.84} & \cellcolor{violet!7}\underline{55.64} & \cellcolor{violet!7}\textbf{62.97} & \cellcolor{violet!7}68.41 & \cellcolor{violet!7}75.60 & \cellcolor{violet!7}\textbf{46.63} & \cellcolor{violet!7}\textbf{49.34} & \cellcolor{violet!7}\textbf{53.26} & \cellcolor{violet!7}\textbf{53.46} & \cellcolor{violet!7}\textbf{64.50}
    \\
    \bottomrule
\end{tabular}
}
\end{table*}

%% file: tables/ablation.tex
\begin{table*}[t]
    \centering
    \caption{Ablation study of components in our framework. All variants are pretrained on \emph{nuScenes} \cite{fong2022panoptic-nuScenes} and fine-tuned on \emph{nuScenes} \cite{fong2022panoptic-nuScenes}, \emph{SemanticKITTI} \cite{behley2019semanticKITTI}, and \emph{Waymo} \cite{sun2020waymoOpen}, respectively. The knowledge is distilled from the ResNet-50. \textbf{C2L:} Camera-to-LiDAR distillation. \textbf{VFM:} Vision foundation models. \textbf{STC:} Superpoint temporal consistency. \textbf{P2S:} Point-to-segment regularization. \textbf{CDP:} Cross-dataset pretraining. \textbf{LP} denotes linear probing with frozen backbones. All mIoU scores are given in percentage (\%). The \textbf{best} and \underline{second best} scores are highlighted in \textbf{bold} and \underline{underline}.}
    \vspace{-0.1cm}
\label{tab:ablation}
\resizebox{\linewidth}{!}{
    \begin{tabular}{p{0.55cm}<{\centering}|p{0.82cm}<{\centering}p{0.82cm}<{\centering}p{0.82cm}<{\centering}p{0.82cm}<{\centering}p{0.82cm}<{\centering}|p{0.82cm}<{\centering}p{0.82cm}<{\centering}p{0.82cm}<{\centering}p{0.82cm}<{\centering}p{0.82cm}<{\centering}p{0.82cm}<{\centering}|p{0.95cm}<{\centering}|p{0.95cm}<{\centering}}
    \toprule
    \multirow{2}{*}{\textbf{\#}} & \multirow{2}{*}{\textbf{C2L}} & \multirow{2}{*}{\textbf{VFM}} & \multirow{2}{*}{\textbf{STC}} & \multirow{2}{*}{\textbf{P2S}} & \multirow{2}{*}{\textbf{CDP}} & \multicolumn{6}{c}{\textbf{nuScenes}} \vline & \textbf{KITTI} & \textbf{Waymo}
    \\
    & & & & & & {\textbf{LP}} & {\textbf{1\%}} & {\textbf{5\%}} & {\textbf{10\%}} &  {\textbf{25\%}} &  {\textbf{Full}} & {\textbf{1\%}} & {\textbf{1\%}}
    \\\midrule\midrule
    (a) & \cellcolor{yellow!8}\checkmark & \cellcolor{yellow!8}& \cellcolor{yellow!8}& \cellcolor{yellow!8}& \cellcolor{yellow!8}& \cellcolor{yellow!8}38.80 & \cellcolor{yellow!8}38.30 & \cellcolor{yellow!8}52.49 & \cellcolor{yellow!8}59.84 & \cellcolor{yellow!8}66.91 & \cellcolor{yellow!8}74.79 & \cellcolor{yellow!8}44.60 & \cellcolor{yellow!8}47.12
    \\\midrule
    (b) & \checkmark & & \checkmark & & & 40.45 & 41.62 & 54.67 & 60.48 & 67.61 & 75.30 & 45.38 & 48.08
    \\
    (c) & \checkmark & \checkmark & & & & 43.00 & 44.02 & 53.03 & 60.84 & 67.38 & 75.21 & 45.72 & 48.75
    \\
    (d) & \checkmark & \checkmark & \checkmark & & & 44.01 & 44.78 & 55.36 & 61.99 & 67.70 & 75.00 & 46.49 & 49.15
    \\
    (e) & \checkmark & \checkmark &  & \checkmark & & 43.35 & 44.25 & 53.69 & 61.11 & 67.42 & 75.44 & 46.07 & 48.82
    \\
    (f) & \checkmark & \checkmark & \checkmark & \checkmark &  & \underline{44.95} & \underline{45.84} & \underline{55.64} & \underline{62.97} & \underline{68.41} & \underline{75.60} & \underline{46.63} & \underline{49.34}
    \\\midrule
    (g) & \cellcolor{violet!7}\checkmark & \cellcolor{violet!7}\checkmark & \cellcolor{violet!7}\checkmark & \cellcolor{violet!7}\checkmark & \cellcolor{violet!7}\checkmark & \cellcolor{violet!7}\textbf{46.13} & \cellcolor{violet!7}\textbf{47.08} & \cellcolor{violet!7}\textbf{56.90} & \cellcolor{violet!7}\textbf{63.74} & \cellcolor{violet!7}\textbf{69.34} & \cellcolor{violet!7}\textbf{76.03} & \cellcolor{violet!7}\textbf{49.55} & \cellcolor{violet!7}\textbf{50.29}
    \\\bottomrule
\end{tabular}
}
\vspace{-0.1cm}
\end{table*}

%% file: tables/sensors.tex
\begin{table*}[t]
    \centering
    \caption{Ablation study on the possible sensor misalignment between the LiDAR and camera sensors. The perturbation is randomly generated and inserted during the pretraining stage. All mIoU scores are given in percentage (\%).}
    \vspace{-0.1cm}
\label{tab:sensors}
\resizebox{\linewidth}{!}{
    \begin{tabular}{r|ccccc|ccccc|ccccc}
    \toprule
    \multirow{2}{*}{\textbf{Method}} & \multicolumn{5}{c}{\textbf{1\% Misalignment}} \vline & \multicolumn{5}{c}{\textbf{5\% Misalignment}} \vline & \multicolumn{5}{c}{\textbf{10\% Misalignment}}
    \\
    & \textbf{CL} & \textbf{1\%} & \textbf{5\%} & \textbf{10\%} & \textbf{25\%} & \textbf{CL} & \textbf{1\%} & \textbf{5\%} & \textbf{10\%} & \textbf{25\%} & \textbf{CL} & \textbf{1\%} & \textbf{5\%} & \textbf{10\%} & \textbf{25\%}
    \\\midrule\midrule
    PPKT \cite{liu2021ppkt} & 35.90 & 34.94 & 51.11 & 58.54 & 65.01 & 35.90 & 33.69 & 51.40 & 58.00 & 64.11 & 35.90 & 33.35 & 50.98 & 57.84 & 63.52
    \\
    SLidR \cite{sautier2022slidr} & 38.80 & 37.92 & 53.08 & 59.89 & 66.90 & 38.80 & 38.00 & 52.36 & 60.01 & 64.10 & 38.80 & 37.30 & 51.11 & 58.50 & 64.50
    \\\midrule
    \cellcolor{violet!7}\textcolor{violet!66}{\textbf{Ours}} & \cellcolor{violet!7}\textbf{45.95} & \cellcolor{violet!7}\textbf{45.23} & \cellcolor{violet!7}\textbf{55.71} & \cellcolor{violet!7}\textbf{62.62} & \cellcolor{violet!7}\textbf{68.13} & \cellcolor{violet!7}\textbf{45.95} & \cellcolor{violet!7}\textbf{45.66} & \cellcolor{violet!7}\textbf{55.42} & \cellcolor{violet!7}\textbf{62.77} & \cellcolor{violet!7}\textbf{68.01} & \cellcolor{violet!7}\textbf{45.95} & \cellcolor{violet!7}\textbf{44.80} & \cellcolor{violet!7}\textbf{54.45} & \cellcolor{violet!7}\textbf{61.80} & \cellcolor{violet!7}\textbf{68.29}
    \\
    \bottomrule
\end{tabular}
}
\vspace{-0.1cm}
\end{table*}

%% file: tables/data_sources.tex
\begin{wraptable}{r}{0.58\textwidth}
\centering
\vspace{-0.5cm}
    \centering
    \caption{Ablation study on the utilization of multiple data sources during pretraining. The knowledge is distilled from the ViT-B. \textbf{N}, \textbf{K}, and \textbf{W} denote pretraining on \emph{nuScenes} \cite{fong2022panoptic-nuScenes}, \emph{SemanticKITTI} \cite{behley2019semanticKITTI}, and \emph{Waymo Open} \cite{sun2020waymoOpen}, respectively. \textbf{LP} denotes linear probing with frozen backbones. All mIoU scores are given in percentage (\%). The \textbf{best} and \underline{second best} scores are highlighted in \textbf{bold} and \underline{underline}.}
    \vspace{-0.1cm}
\label{tab:sources}
\resizebox{\linewidth}{!}{
    \begin{tabular}{ccc|cc|cc|cc}
    \toprule
    \multirow{2}{*}{\textbf{N}} & \multirow{2}{*}{\textbf{K}} & \multirow{2}{*}{\textbf{W}} & \multicolumn{2}{c}{\textbf{nuScenes}} \vline & \multicolumn{2}{c}{\textbf{KITTI}} \vline & \multicolumn{2}{c}{\textbf{Waymo}}
    \\
    & & & {\textbf{LP}} & {\textbf{1\%}} & {\textbf{LP}} & {\textbf{1\%}} &  {\textbf{LP}} &  {\textbf{1\%}} 
    \\\midrule\midrule
    \cellcolor{yellow!8} & \cellcolor{yellow!8} & \cellcolor{yellow!8} & \cellcolor{yellow!8}8.10 & \cellcolor{yellow!8}30.30 & \cellcolor{yellow!8}7.45 & \cellcolor{yellow!8}39.50 & \cellcolor{yellow!8}6.45 & \cellcolor{yellow!8}39.41
    \\\midrule
    \checkmark & & & 46.59 & 45.98 & 29.25 & 47.24 & 32.42 & 48.91
    \\
    & \checkmark & & 18.18 & 38.86 & 33.46 & 47.52 & 29.36 & 45.96 
    \\
    & & \checkmark & 25.74 & 42.69 & 33.21 & 47.46 & 34.65 & 49.30 
    \\\midrule
    \checkmark & \checkmark & & 47.19 & 47.80 & 34.09 & 49.12 & 31.34 & 50.00 
    \\
    \checkmark & & \checkmark & \underline{47.42} & \underline{47.94} & 33.16 & 48.61 & \underline{35.84} & 50.81 
    \\
    & \checkmark & \checkmark & 31.08 & 44.77 & \underline{34.67} & \underline{50.16} & 35.49 & \underline{50.96} 
    \\\midrule
    \cellcolor{violet!7}\checkmark & \cellcolor{violet!7}\checkmark & \cellcolor{violet!7}\checkmark & \cellcolor{violet!7}\textbf{47.84} & \cellcolor{violet!7}\textbf{48.37} & \cellcolor{violet!7}\textbf{35.77} & \cellcolor{violet!7}\textbf{50.68} & \cellcolor{violet!7}\textbf{36.12} & \cellcolor{violet!7}\textbf{51.52} 
    \\\bottomrule
\end{tabular}
}
\vspace{-0.5cm}
\end{wraptable}

%% file: sections/6_conclusion.tex
\section{Conclusion}
\label{sec:conclusion}
In this paper, we introduced \textbf{LargeAD}, a scalable and generalizable framework designed for large-scale pretraining across diverse LiDAR datasets. Our approach leverages vision foundation models (VFMs) to generate semantically enriched superpixels, aligning 2D image features with LiDAR point clouds for improved representation learning. By integrating a VFM-assisted contrastive learning objective, superpoint temporal consistency, and multi-source data pretraining, our framework achieves state-of-the-art performance on multiple 3D scene understanding tasks, including LiDAR-based semantic segmentation and 3D object detection. Extensive experiments conducted on eleven diverse datasets highlight the effectiveness of our framework in both in-domain and out-of-domain scenarios. LargeAD excels not only in downstream generalization but also demonstrates strong robustness under out-of-distribution conditions. The ablation study further validates the importance of our design choices, showcasing the significant impact of incorporating multiple datasets during pretraining and the benefits of each individual component within our framework. Our results underline the potential of LargeAD to support real-world autonomous driving by enabling versatile and resilient models that adapt to diverse sensor configurations and driving environments. In future work, we aim to extend our approach to incorporate additional sensor modalities, such as radar and thermal imaging, further broadening the scope of cross-modal pretraining for autonomous systems.